\documentclass{article}

\usepackage{microtype}
\usepackage{graphicx}
\usepackage{subfigure}
\usepackage{booktabs} 

\usepackage{hyperref}

\usepackage[accepted]{mlsys2020}

\usepackage{graphicx}
\usepackage{enumitem}
\usepackage[textsize=tiny,disable]{todonotes}
\usepackage{textcomp}
\usepackage{lscape}
\usepackage{math}
\usepackage[symbol]{footmisc}
\usepackage{tikz}
\usepackage[makeroom]{cancel}

\mlsystitlerunning{Accounting for Variance
                   in Machine Learning Benchmarks}

\begin{document}

\twocolumn[
\mlsystitle{Accounting for Variance in Machine Learning Benchmarks}



\begin{mlsysauthorlist}
\mlsysauthor{Xavier Bouthillier}{mila,udem}
\mlsysauthor{Pierre Delaunay}{independent}
\mlsysauthor{Mirko Bronzi}{mila}
\mlsysauthor{Assya Trofimov}{mila,udem,iric}
\mlsysauthor{Brennan Nichyporuk}{mila,cim,mcgill}
\mlsysauthor{Justin Szeto}{mila,cim,mcgill}
\mlsysauthor{Naz Sepah}{mila,cim,mcgill}
\mlsysauthor{Edward Raff}{bah,mary}
\mlsysauthor{Kanika Madan}{mila,udem}
\mlsysauthor{Vikram Voleti}{mila,udem}
\mlsysauthor{Samira Ebrahimi Kahou}{mila,mcgill,ets,cifar}
\mlsysauthor{Vincent Michalski}{mila,udem}
\mlsysauthor{Dmitriy Serdyuk}{mila,udem}
\mlsysauthor{Tal Arbel}{mila,cim,mcgill,cifar}
\mlsysauthor{Chris Pal}{mila,poly,element}
\mlsysauthor{Gaël Varoquaux}{mila,mcgill,inria}
\mlsysauthor{Pascal Vincent}{mila,udem,cifar}
\end{mlsysauthorlist}

\mlsysaffiliation{mila}{Mila, Montréal, Canada}
\mlsysaffiliation{independent}{Independent}
\mlsysaffiliation{udem}{Université de Montréal, Montréal, Canada}
\mlsysaffiliation{iric}{IRIC}
\mlsysaffiliation{cim}{Centre for Intelligent Machines}
\mlsysaffiliation{mcgill}{McGill University, Montréal, Canada}
\mlsysaffiliation{bah}{Booz Allen Hamilton}
\mlsysaffiliation{mary}{University of Maryland, Baltimore County}
\mlsysaffiliation{poly}{Polytechnique Montréal, Montréal, Canada}
\mlsysaffiliation{element}{ElementAI}
\mlsysaffiliation{inria}{Inria, Saclay, France}
\mlsysaffiliation{cifar}{CIFAR}
\mlsysaffiliation{ets}{École de technologie supérieure}

\mlsyscorrespondingauthor{Xavier Bouthillier}{xavier.bouthillier@umontreal.ca}

\mlsyskeywords{}

\vskip 0.3in

\begin{abstract}
Strong empirical evidence that one machine-learning algorithm 
$A$ outperforms another one $B$ ideally calls for multiple trials
optimizing the learning pipeline over sources of
variation such as data sampling, augmentation, parameter initialization,
and hyperparameters choices. This is prohibitively expensive, and
corners are cut to reach conclusions. We model the whole benchmarking
process, revealing that variance due to
data sampling, parameter initialization and hyperparameter choice impact
markedly the results.
We analyze the predominant comparison methods used today in the light of
this variance. We show a
counter-intuitive result that adding more sources of variation to an
imperfect estimator approaches better the ideal estimator at a
$51\times$ reduction in compute cost. Building on these results, we study
the error rate of detecting improvements, on five
different deep-learning tasks/architectures. This study leads us to
propose recommendations for performance comparisons. 
\end{abstract}
]%



\printAffiliationsAndNotice{}  

\section{Introduction: trustworthy benchmarks account for fluctuations}


Machine learning increasingly relies upon empirical evidence to validate
publications or  efficacy.
The value of a new method or algorithm $A$ is often
established by empirical benchmarks comparing it to prior work. 
Although such benchmarks are built on quantitative measures of
performance, uncontrolled factors can impact these measures and dominate
the meaningful difference between the methods.
In particular, recent studies have shown that loose choices of
hyper-parameters lead to non-reproducible benchmarks and unfair comparisons \cite{Raff2019_quantify_repro,Raff2020c,NIPS2018_7350,henderson2018deep,
kadlec2017knowledge,
melis2018state,
pmlr-v97-bouthillier19a,
reimers-gurevych-2017-reporting,
gorman-bedrick-2019-need}. 
Properly accounting for these factors may go as far as changing the conclusions for the comparison, as shown for recommender systems \cite{Dacrema2019}, 
neural architecture pruning \cite{mlsys2020_73}, and metric 
learning \cite{Musgrave2020}.

The steady increase in complexity --e.g.\ neural-network depth--
and number of hyper-parameters of 
learning pipelines increases computational costs of models, making brute-force approaches prohibitive.
Indeed, robust conclusions on comparative performance of models $A$ and $B$ would require multiple training of the full learning pipelines, including hyper-parameter optimization and random seeding.
Unfortunately, since the computational budget of most researchers can
afford only a small number of model fits \cite{bouthillier:hal-02447823},
many sources of variances are not probed via repeated experiments.
Rather, sampling several model initializations is often considered to
give enough evidence.
As we will show, there are other, larger, sources of uncontrolled
variation and the risk is that conclusions are driven by differences due to
arbitrary factors, such as data order, rather than model improvements.

The seminal work of \citet{dietterich1998approximate} studied statistical tests 
for comparison of supervised classification learning algorithms focusing on 
variance due to data sampling. Following works 
\cite{NIPS1999_1661, bouckaert2004evaluating} perpetuated this focus,
including a series of work in NLP 
\cite{riezler2005-significance62,kirkpatrick2012-investigation75,sogaard2014-value34} 
which ignored variance extrinsic to data sampling. Most of these works recommended the use of
paired tests to mitigate the issue of extrinsic sources of variation, but
\citet{hothorn2005design} then proposed a theoretical framework
encompassing all sources of variation. This framework addressed the issue
of extrinsic sources of variation
by marginalizing all of them, including the hyper-parameter optimization
process. These prior works need to be confronted to the current practice
in machine learning, in particular deep learning, where \emph{1)} the
machine-learning pipelines has a large number of hyper-parameters,
including to define the architecture, set by uncontrolled procedures, sometimes
manually, \emph{2)} the
cost of fitting a model is so high that train/validation/test splits are
used instead of cross-validation, or nested cross-validation that
encompasses hyper-parameter optimization \citep{bouthillier:hal-02447823}.

In \textbf{Section~\ref{sec:variance}}, we study the different source of 
variation of a benchmark, to outline which factors contribute
markedly to uncontrolled fluctuations in the measured performance.
\textbf{Section~\ref{sec:estimation}} discusses estimation
the performance of a pipeline and its uncontrolled variations with a
limited budget. In particular we discuss this estimation when
hyper-parameter optimization is run only once.
Recent studies emphasized that
model comparisons with uncontrolled hyper-parameter optimization
is a burning issue
\cite{NIPS2018_7350,henderson2018deep,kadlec2017knowledge,melis2018state,pmlr-v97-bouthillier19a};
here we frame it in a statistical context, with explicit bias and
variance to measure the loss of reliability that it incurs. In
\textbf{Section~\ref{sec:error-rates}}, we discuss criterion using
these estimates to
conclude on whether to accept algorithm $A$ as a meaningful improvement
over algorithm $B$, and the error rates that they incur in the face of
noise.

Based on our results, we issue in \textbf{Section~\ref{sec:recommendations}} the following recommendations: 
\begin{enumerate}[leftmargin=3ex,noitemsep,topsep=0pt,parsep=0pt,partopsep=0pt,label=\textbf{\arabic*)}]
  \item
As many sources of variation as possible should be randomized whenever possible. These include weight initialization, data sampling, random data augmentation and the whole hyperparameter optimization. 
This helps decreasing the standard error of the average performance estimation, 
enhancing precision of benchmarks.
  \item Deciding of whether the benchmarks give evidence that one
algorithm outperforms another should not build solely on comparing
average performance but account for variance. We propose a simple
decision criterion based on requiring a high-enough probability that in 
one run an algorithm outperforms another.
  \item
Resampling techniques such as out-of-bootstrap should be favored instead of fixed held-out test sets to
improve capacity of detecting small improvements.
\end{enumerate}
Before concluding, we outline a few additional considerations for benchmarking in
\textbf{Section~\ref{sec:additional_considerations}}.

\section{%
    The variance in ML benchmarks
}
\label{sec:variance}

Machine-learning benchmarks run a complete learning pipeline on a finite
dataset to estimate its performance.
This performance value should be considered the realization of a random
variable. Indeed the
dataset is itself a random sample from the full data distribution. In
addition, a typical
learning pipeline has additional sources of uncontrolled fluctuations, as we will
highlight below. A proper evaluation and comparison between
pipelines should thus account for the \emph{distributions} of such metrics.


\subsection{A model of the benchmarking process that includes
hyperparameter tuning}
\label{sec:learning_framework}

Here we extend the formalism of \citet{hothorn2005design} to model
the different sources of variation in a machine-learning pipeline and
that impact performance measures.
In particular, we go beyond prior works by accounting for the choice of
hyperparameters in a probabilistic model of the whole experimental
benchmark. Indeed, choosing good hyperparameters --including details of a neural
architecture-- is crucial to the performance of a pipeline. Yet these
hyperparameters come with uncontrolled noise, whether they are set
manually or with an automated procedure.

\paragraph{The training procedure}
We consider here the familiar setting of
supervised learning on i.i.d. data (and will use classification in our
experiments) but this can easily be adapted to other machine learning settings.
Suppose we have access to a dataset $S=\{(x_1, y_1), \ldots, (x_n, y_n)\}$ containing
$n$ examples of (input, target) pairs.
These pairs are i.i.d. and sampled from an unknown data
distribution $\mathcal{D}$, i.e. $S \sim \mathcal{D}^n$.
The goal of a learning pipeline is to find a function $h \in \mathcal{H}$ that will have good prediction performance in
expectation over $\mathcal{D}$, 
as evaluated by a metric of interest $e$. More precisely, in supervised
learning, $e(h(x),y)$ is a
measure of how far a prediction $h(x)$ lies from the
target $y$ associated to the input $x$
(e.g., classification error).
The goal is to find a predictor $h$ that minimizes the \emph{expected
risk} $R_e(h, \mathcal{D}) = \mathbb{E}_{(x,y)\sim \mathcal{D}}[e(h(x),y)]$, but since we
have access only to finite datasets, all we can ever measure is an \emph{empirical risk}
$\hat{R}_e(h, S) = \frac{1}{|S|} \sum_{(x,y) \in S}  e(h(x),y)$.
%
In practice training with a training set $S^t$ consists in finding a
function (hypothesis) $h\in\mathcal{H}$ that minimizes a trade-off
between a data-fit term --typically the empirical risk of a
differentiable surrogate loss $e'$-- with a regularization $\Omega(h, \lambda)$ that induces a
preference over hypothesis functions:
\begin{equation}
  \opt(S^t, \lambda)
  \approx \arg\min_{h\in\mathcal{H}} \hat{R}_{e'}(h, S^t)
	+ \Omega(h, \lambda),
\end{equation}
where $\lambda$ represents the set of hyperparameters: regularization
coefficients (s.a. strength of weight decay or the ridge penalty), 
architectural hyperparameters affecting $\hypo$, optimizer-specific ones such as the learning rate, etc\ldots Note that $\opt$ is a
random variable whose value will depend also on other additional random variables that we
shall collectively denote $\xi_O$, sampled to determine parameter
initialization, data augmentation, example ordering, etc.\footnote{If stochastic data augmentation is used, then
optimization procedure $\mathrm{Opt}$ for a given training set $S^t$ has to be
changed to  an expectation over $\tilde{S^t} \sim
P^\mathrm{aug}(\tilde{S^t}|{S^t}; \lambda_\mathrm{aug})$ where
$P^\mathrm{aug}$ is the data augmentation distribution.
This adds additional stochasticity to the optimization, as we will optimize this
through samples from  $P^\mathrm{aug}$ obtained with a random number generator.}.
\paragraph{Hyperparameter Optimization}
The training procedure builds a predictor given a training set $S^t$. But
since it requires specifying hyperparameters
$\lambda$, a complete learning pipeline has to tune all of these. A complete pipeline will involve a hyper-parameter optimization
procedure, which will strive to find a value of $\lambda$ that minimizes objective
\begin{equation}\label{eq:hpo_loss}
r(\lambda) = \mathbb{E}_{(S^t, S^v)\sim
\mathrm{sp}(S^{tv})} \left[ \hat{R}_e\left(\opt(S^t,\lambda), S^v\right) \right]
\end{equation}
where $\mathrm{sp}(S^{tv})$ is a distribution of random splits of the data
set $S^{tv}$ between training and validation subsets $S^t, S^v$.
Ideally, hyperparameter optimization would be applied
over random dataset samples from the true distribution $\dist$, 
but in practice the learning pipeline only has access to $S^{tv}$, hence the
expectation over dataset splits. An ideal hyper-parameter optimization would yield
$\lambda^*(S^{tv}) = \arg\min_\lambda r(\lambda)$.
A concrete hyperparameter optimization
algorithm $\mathrm{HOpt}$ will however use an average over a small number
of train-validation splits (or just 1), and a limited training budget, yielding 
${\lambdaopt(S^{tv}) = \Hopt(S^{tv}) \approx \lambda^*(S^{tv})}$.
We denoted earlier the sources of random variations in $\opt$ as $\xi_O$.
Likewise, we will denote the sources of variation inherent to $\Hopt$ as $\xi_H$.
These encompass the sources of variance related to the procedure to 
optimize hyperparameters $\mathrm{HOpt}$, whether it is manual or
a search procedure which has its arbitrary choices such as the splitting and random exploration.

After hyperparameters have been tuned, it is often customary to retrain the
predictor using the full data $S^{tv}$. The complete learning pipeline
$\mathcal{P}$ will finally
return a single predictor:
\begin{equation} \label{eq:hopt_s}
  \hopt (S^{tv}) = \mathcal{P}(S^{tv}) = \opt(S^{tv}, \Hopt(S^{tv}))
\end{equation}
  Recall that $\hopt (S^{tv})$ is the result of $\opt$ which is not deterministic, as it is affected by
arbitrary choices $\xi_O$ in the training of the model (random weight
initialization, data ordering...) and now additionally $\xi_H$ in the
  hyperparameter optimization. 
  We will use $\xi$ to denote the set of all random variations sources in the learning
  pipeline, $\xi = \xi_H \cup \xi_O$. 
  Thus $\xi$ captures all sources of variation in the learning pipeline from data $S^{tv}$, that are not configurable with $\lambda$.

\paragraph{The performance measure}

The full learning procedure $\mathcal{P}$ described above yields a model $\hopt$. We now must define
a metric that we can use to evaluate the performance of this model with statistical tests.
For simplicity, we
will use the same evaluation metric $e$ on which we based hyperparameter
optimization.
The expected risk obtained 
by applying the full learning pipeline  $\mathcal{P}$ to datasets 
$S^{tv}\sim \mathcal{D}^n$ of size $n$ is:
\begin{equation} \label{eq:expected_risk}
  R_\mathcal{P}(\mathcal{D}, n)
  = \mathbb{E}_{S^{tv} \sim \mathcal{D}^n} \left[  R_e(\hopt (S^{tv}),  \mathcal{D}) \right] 
\end{equation}
where the expectation is also over the random sources $\xi$
that affect the learning procedure (initialization, ordering,
data-augmentation) and hyperparameter optimization.

As we only have access to a single finite dataset $S$, the performance of the learning pipeline can be evaluated as the
following expectation over splits:
\begin{align} \label{eq:empirical_average_risk}
  \mu=\hat{R}_\mathcal{P}(S,\, &n,n') = \nonumber\\
  & \mathbb{E}_{(S^{tv}, S^o) \sim \mathrm{sp}_{n,n'}(S)}
      \left[  \hat{R}_e(\hopt (S^{tv}), S^o) \right]
\end{align}
where $\mathrm{sp}$ is a distribution of random splits or bootstrap
resampling of the data set $S$ that yield sets $S^{tv}$ (train+valid) of size $n$ and
$S^o$ (test) of size $n'$.
We denote as $\sigma^2$ the corresponding variance of $\hat{R}_e(\hopt (S^{tv}), S^o)$.
The performance measures vary not only depending on how the data was
split, but also on all other random factors affecting the learning
procedure $(\xi_O)$ and hyperparameters optimization $(\xi_H)$.

\subsection{Empirical evaluation of variance in benchmarks}
\label{sec:exps-var}

\begin{figure*}
  \includegraphics[width=\textwidth]{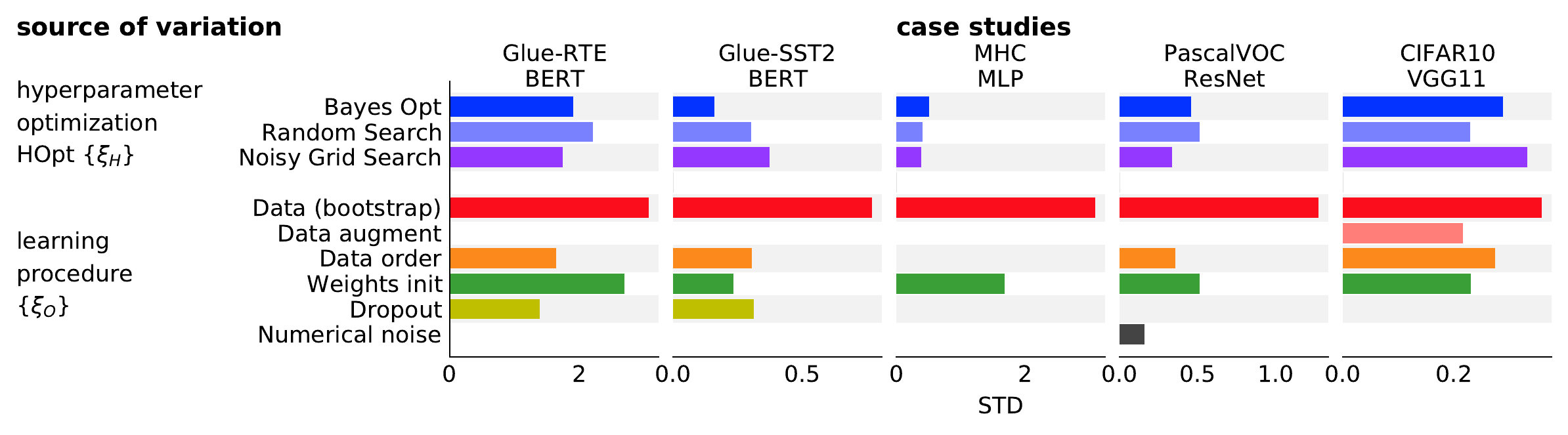}%
    \caption{\textbf{Different sources of variation of the measured
performance}: across our different case studies, as a fraction of the
variance induced by bootstrapping the data. For hyperparameter
optimization, we studied several algorithms.}
  \label{fig:variances}
\end{figure*}

We conducted thorough experiments to probe the different sources
of variance in machine learning benchmarks.

\paragraph{Cases studied}
We selected i) the CIFAR10 \citep{krizhevsky2009learning}
image classification with VGG11 \citep{simonyan2014very},
ii) PascalVOC \citep{pascal-voc-2012} image segmentation  using an FCN \cite{long2014fully} with a ResNet18 \cite{he2015deep} backbone pretrained on imagenet \cite{imagenet_cvpr09},
iii-iv) Glue \cite{wang2019glue} SST-2 \cite{socher2013recursive} and RTE \cite{bentivogli2009fifth} tasks with BERT \cite{devlin2018bert}
and v) peptide to major histocompatibility class I (MHC I) binding predictions with a shallow MLP.
All details on default hyperparameters used and the computational
environments --which used $\sim 8$ GPU years-- can be found in Appendix~\ref{sec:case-studies-appendix}.


\paragraph{Variance in the learning procedure: $\xi_O$}
\label{sec:exp_var_o}
For the sources of variance from the learning procedure ($\xi_O$), we
identified: i) the data sampling, ii) data augmentation procedures, iii)
model initialization, iv) dropout, and v) data visit order in stochastic gradient descent.
We model the data-sampling variance as resulting from training the model on a finite dataset $S$ of size $n$, sampled from an unknown true distribution. 
$S \sim \dist^n$ is thus a random
variable, the standard source of variance considered in statistical learning.
Since we have a single finite dataset in practice, we evaluate this variance by repeatedly
generating a train set from 
bootstrap replicates of the data and measuring the out-of-bootstrap
error \cite{hothorn2005design}%
\footnote{The more common alternative in machine learning is to use cross-validation, but the latter
is less amenable to various sample sizes. Bootstrapping is discussed in more detail in Appendix ~\ref{sec:bootstrap-appendix}.}. 

We first fixed hyperparameters to pre-selected reasonable
choices\footnote{This choice is detailed in Appendix~\ref{sec:case-studies-appendix}.}.
Then, iteratively for each sources of variance, we randomized the seeds 200 times, while keeping all other sources fixed to initial values. 
Moreover, we measured the numerical noise with 200 training runs with all fixed seeds.

Figure~\ref{fig:variances} presents the individual variances due to sources from within the learning algorithms. 
Bootstrapping data stands out as the most important
source of variance.
In contrast, model initialization generally is less than 50\% 
of the variance of bootstrap, on par with the visit order of stochastic gradient
descent. Note that these different contributions to the variance
are not independent, the total variance cannot be obtained by simply
adding them up. 

\begin{figure}[t]
  \includegraphics[width=\columnwidth]{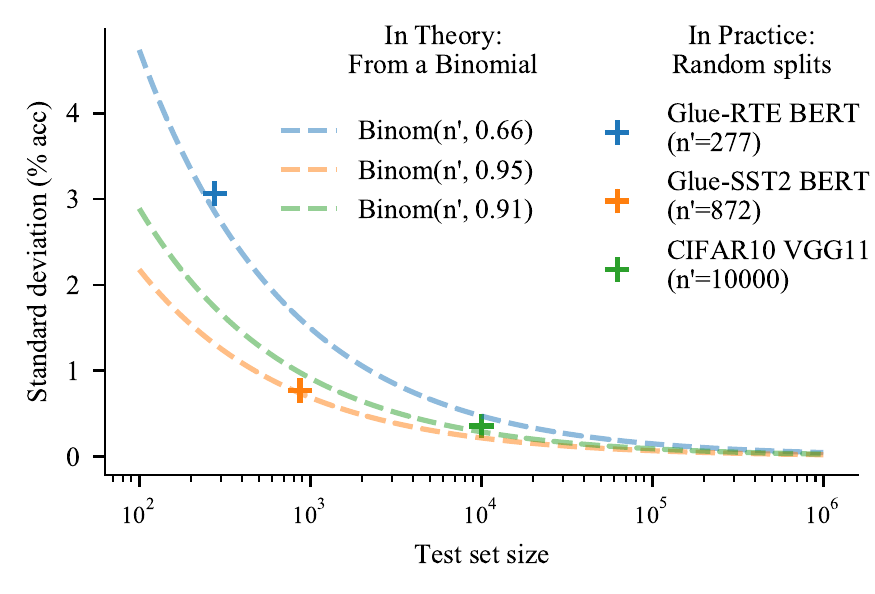}
  \vspace{-2em}
  \caption{\textbf{Error due to data sampling:} The dotted lines show the
standard deviation given by a binomial-distribution model of the accuracy
measure; the crosses report the standard deviation observed when
bootstrapping the data in our case studies, showing that the model is a reasonable.
    \label{fig:binomial_model}
  }
\end{figure}

For classification, a simple binomial can be used to
model the sampling noise in the measure of the prediction accuracy of
a trained pipeline on the test set. Indeed, if the pipeline has a chance
$\tau$ of giving the wrong answer on a sample, makes i.i.d. errors,
and is measured on $n$ samples, the observed measure follows a binomial
distribution of location parameter $\tau$ with $n$ degrees of freedom. If
errors are correlated, not i.i.d., the degrees of freedom are smaller and
the distribution is wider.
Figure \ref{fig:binomial_model} compares standard deviations of the
performance measure given by this simple binomial model to those
observed when bootstrapping the data on the three classification case
studies. The match between the model and the empirical results suggest
that the variance due to data sampling is well explained by the limited
statistical power in the test set to estimate the true performance.

\paragraph{Variance induced by hyperparameter optimization:
$\xi_H$}
\label{sec:exp_var_hpo}
To study the $\xi_H$ sources of variation, we chose three of the most
popular hyperparameter optimization methods: i) random search, ii) grid
search, and iii) Bayesian optimization.
While grid-search in itself has no random parameters, the
specific choice of the parameter range is arbitrary and can be an uncontrolled source of variance (e.g., does the grid size step by powers of 2, 10, or increments of 0.25 or 0.5). 
We study this variance with a \textit{noisy grid search}, perturbing slightly the parameter ranges
(details in Appendix~\ref{sec:hpo-methods-appendix}).

For each of these tuning methods, we held all $\xi_O$ fixed to random values and executed 20 independent hyperparameter optimization procedures up to a budget of 200 trials.
This way, all the observed variance across the hyperparameter optimization procedures is strictly due to $\xi_H$. 
We were careful to design the search space so that it covers the optimal hyperparameter values (as stated in original studies) while being large enough to cover suboptimal values as well.

Results in figure~\ref{fig:variances} show that hyperparameter
choice induces a sizable amount of variance, not negligible in comparison
to the other factors. 
The full optimization curves of the 320 HPO procedures are presented in 
Appendix~\ref{sec:hpo-methods-results}.
The three hyperparameter optimization methods 
induce on average as much variance as the commonly studied weights initialization.
These results motivate further investigation
the cost of ignoring
 the variance due to hyperparameter optimization.

%

%
%


\begin{figure}[t]
  \includegraphics[width=\columnwidth]{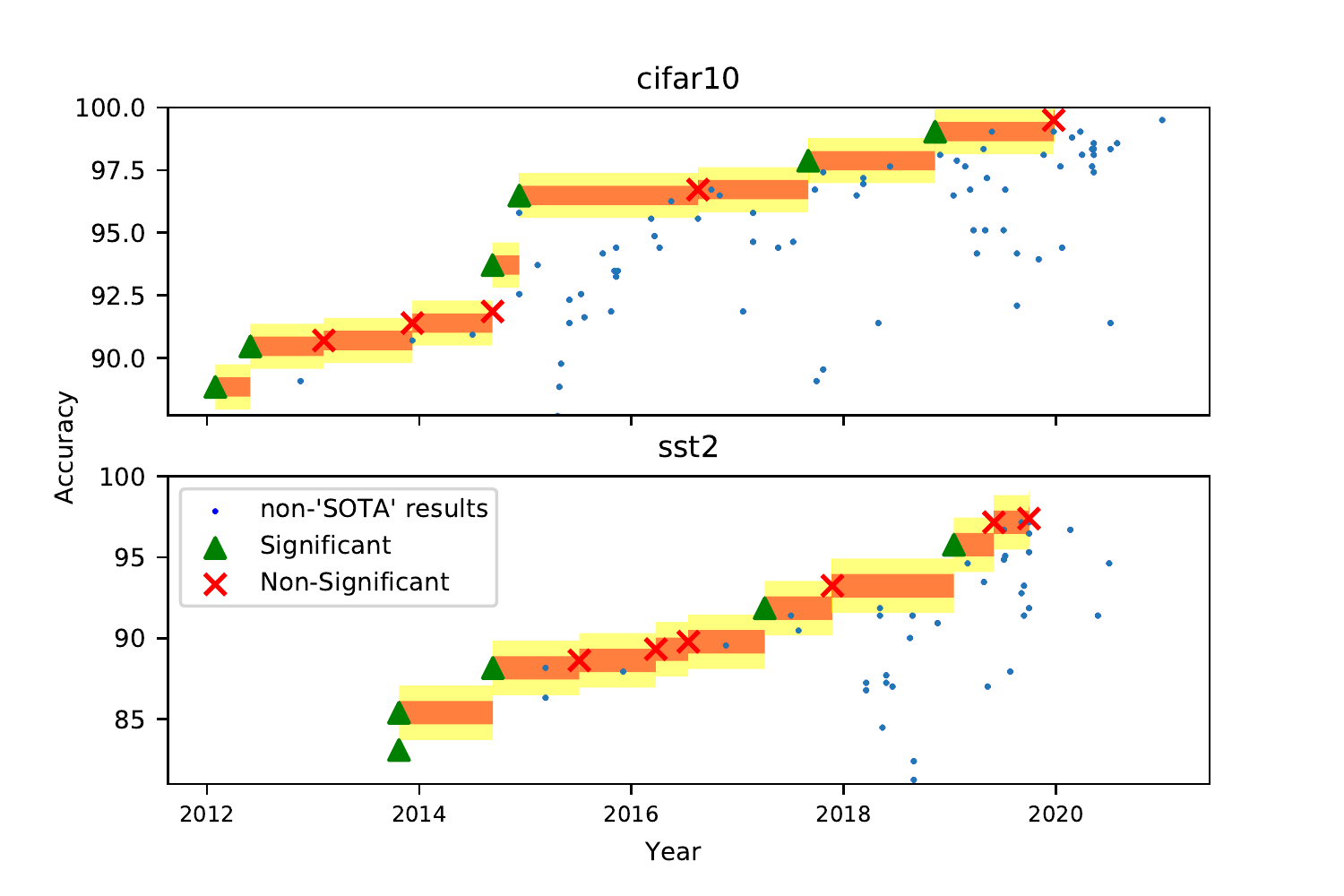}
  \vspace{-2em}
  \caption{\textbf{Published improvements compared to benchmark variance}
    The dots give the performance of publications, function of year, as
    reported on \url{paperswithcode.com}; red band shows our estimated $\sigma$, and the yellow band 
    the resulting significance threshold. Green marks are results likely significant
    compared to prior 'State of the Art', and red 
    "$\times$" 
    appear non-significant.}
    \label{fig:paperswithcode}%
\end{figure}

\paragraph{The bigger picture: Variance matters}

For a given case study, the total variance due to arbitrary choices and
sampling noise revealed by our study can be put in perspective with the
published improvements in the state-of-the-art. Figure 
\ref{fig:paperswithcode} shows that this variance is on the order of
magnitude of the individual increments. In other words, the variance is
not small compared to the differences between pipelines. It 
must be accounted for when benchmarking pipelines.

\section{Accounting for variance to reliably estimate performance $\expr$}
\label{sec:estimation}

This section contains 
1) an explanation of the counter intuitive result that accounting for more sources
of variation reduces the standard error for an estimator of $\expr$
and 2) an empirical measure of
the degradation of expected empirical risk estimation 
due to neglecting
$\Hopt$ variance. 

We will now consider different \emph{estimators} of the average
performance $\mu = {\hat{R}_{\mathcal{P}}(S,n,n^{\prime})}$
from Equation~\ref{eq:empirical_average_risk}. Such estimators will use, in place of the expectation of Equation~\ref{eq:empirical_average_risk}, an
empirical average over $k$ (train+test) splits, which we will denote $\hmuk$ and $\hsigk^2$ the corresponding empirical variance.
We will make an important distinction between an estimator which encompasses all sources of variation,
the \emph{ideal estimator} $\hmuk$, and one which accounts only for a portion of these sources, 
the \emph{biased estimator} $\tmuk$.

But before delving into this,
we will explain why many splits help 
estimating the expected empirical risk 
($\expr$).

\subsection{Multiple data splits for smaller detectable improvements}
\label{sec:fixed-held-out}

The majority of machine-learning benchmarks are built with fixed training and test sets.
The rationale behind this design, is that learning algorithms should be compared
on the same grounds, thus on the same sets of examples for training and testing.
While the rationale is valid, it disregards the fact that the fundamental ground of comparison is
the true distribution from which the sets were sampled.
This finite set is used to compute the expected empirical risk 
($\expr$ Eq~\ref{eq:empirical_average_risk}),
failing to compute the expected risk ($\erisk$ Eq~\ref{eq:expected_risk}) on the
whole distribution. This empirical risk is therefore a noisy measure, it
has some uncertainty because the risk on a particular test set gives limited
information on what would be the risk on new data. This uncertainty due to data
sampling is not small compared to typical improvements or other sources
of variation, as revealed by our study in the previous section. In
particular, figure \ref{fig:binomial_model} suggests that the size of the test set can be a limiting
factor.

When comparing two learning algorithms $A$ and $B$, we estimate their expected empirical
risks $\expr$ with $\hmuk$, a noisy measure. The uncertainty of this
measure is represented by the standard error $\frac{\sigma}{\sqrt{k}}$
under the normal assumption\footnote{Our extensive numerical experiments
show that a normal distribution is well suited for the fluctuations of
the risk--figure \ref{fig:normal}} of $\hat{R}_e$.
This uncertainty is an important aspect of the comparison, for instance
it appears in statistical tests used to draw a conclusion
in the face of a noisy evidence.
For instance, a z-test states that a difference of expected empirical risk between $A$ and $B$ of at
least $z_{0.05}\sqrt{\frac{\sigma_A^2 + \sigma_B^2}{k}}$ must be observed to control false
detections at a rate of 95\%.
In other words, a difference smaller than this value
could be due to noise alone, e.g.\ different sets of random splits may lead to different conclusions.

With $k=1$, algorithms $A$ and $B$ must have a large difference of
performance to support a reliable detection.
In order to detect smaller differences, $k$ must be increased, i.e. $\hmuk$ must be
computed over several data splits.
The estimator $\hmuk$ is computationally expensive however, and most
researchers must instead use a biased estimator $\tmuk$ that does not
probe well all sources of variance. 

\subsection{Bias and variance of estimators depends on whether they
account for all sources of variation}
\label{sec:stderr}

Probing all sources of variation, including hyperparameter optimization, 
is too computationally expensive
for most researchers. However, ignoring the role of hyperparameter optimization
induces a bias in the estimation of the expected empirical risk.
We discuss in this section the expensive, unbiased, ideal estimator of $\hmuk$ and the 
cheap biased estimator of $\tmuk$. We explain as well why accounting for
many sources of variation improves the biased estimator by reducing its
bias.%


\subsubsection{Ideal estimator: sampling multiple HOpt}
\label{sec:ideal-estimator}

The ideal estimator $\hmuk$ takes into account all sources of variation.
For each performance measure $\hat{R}_e$, all $\xi_O$ and $\xi_H$ are randomized, each requiring an independent
hyperparameter optimization procedure.
The detailed procedure is presented in Algorithm~\ref{alg:ideal-estimator}.
For an estimation over $k$ splits with hyperparameter optimization for a budget of $T$ trials,
it requires fitting the learning algorithm a total of $O(k \cdot T)$ times.
The estimator is unbiased, with $\E\left[\hmuk\right]=\mu$.

For a variance of the performance measures $\Var(\hre)=\sigma^2$, we can derive the
variance of the ideal estimator
$\Var(\hmuk)=\frac{\sigma^2}{k}$ by taking
the sum of the variances in $\hmuk = \frac{1}{k}\sum_{i=1}^k\hat{R}_{ei}$.
We see that with $\lim_{k\rightarrow\infty}\Var(\hmuk)=0$. Thus $\hmuk$ is a 
well-behaved unbiased
estimator of $\mu$, as its mean squared error vanishes 
with $k$ infinitely large:
\begin{align}
  \E[(\hmuk - \mu)^2] &= \Var(\hmuk) + (\E[\hmuk] - \mu)^2 \nonumber\\
                      &= \frac{\sigma^2}{k}
\end{align}
Note that $T$ does not appear in these equations. Yet it controls
$\operatorname{HOpt}$'s runtime cost ($T$ trials to determine
$\hat{\lambda^*}$), and thus the variance $\sigma^2$ is a function of $T$.

%
%
\begin{figure}%
\vspace{-1em}
\tikzstyle{every picture}+=[remember picture]%
\begin{minipage}[t]{0.50\linewidth}
 \begin{algorithm}[H]
    \caption{{\tt IdealEst}\\Ideal Estimator $\hmuk,\hsigk$
    }
    \label{alg:ideal-estimator}
  \begin{algorithmic}
   \STATE {\bfseries Input:}
   \STATE \hspace{1em} dataset $S$
   \STATE \hspace{1em} sample size $k$
   \STATE
   \FOR{i in $\{1,\cdots,k\}$}
   \STATE $\xi_O \sim \mbox{RNG}()$
   \STATE $\xi_H \sim \mbox{RNG}()$
   \STATE $S^{tv}, S^{o} \sim \mathrm{sp}(S;\xi_O)$
   \STATE \tikz[baseline]{ \node[anchor=base,inner sep=0em]
	(hopt1) {\color[HTML]{3531FF}
	    $\lambdaopt = \mbox{HOpt}(S^{tv},\xi_O,\xi_H)$
	};}
   \STATE $\hopt = \opt(S^{tv}, \lambdaopt)$
   \STATE $p_i = \hat{R}_e(\hopt, S^o)$
   \ENDFOR
   \STATE {\bfseries Return} $\hmuk=\mbox{mean}(p)$,
   \STATE {\hspace{3em}} $\hsigk=\mbox{std}(p)$
  \end{algorithmic}
  \end{algorithm}
\end{minipage}
\hfill
\begin{minipage}[t]{0.49\linewidth}
 \begin{algorithm}[H]
    \caption{{\tt FixHOptEst}\\Biased Estimator $\tmuk,\tsigk$
    }
    \label{alg:biased-estimator}
  \begin{algorithmic}
   \STATE {\bfseries Input:}
   \STATE \hspace{1em} dataset $S$
   \STATE \hspace{1em} sample size $k$
   \STATE
   \STATE $\xi_O \sim \mbox{RNG}()$
   \STATE $\xi_H \sim \mbox{RNG}()$
   \STATE $S^{tv}, S^{o} \sim \mathrm{sp}(S;\xi_O)$
   \STATE \tikz[baseline]{ \node[anchor=base,inner sep=0em]
	(hopt2) {\color[HTML]{3531FF}
	    $\hat{\lambda}^*= \mbox{HOpt}(S^{tv},\xi_O,\xi_H)$
	};}
   \FOR{i in $\{1,\cdots,k\}$}
   \STATE $\xi_O \sim \mbox{RNG}()$
   \STATE $S^{tv}, S^{o} \sim \mathrm{sp}(S;\xi_O)$
   \STATE $\hopt = \opt(S^{tv}, \lambdaopt)$
   \STATE $p_i = \hat{R}_e(\hopt, S^o)$
   \ENDFOR
   \STATE {\bfseries Return} $\tmuk=\mbox{mean}(p)$,
   \STATE {\hspace{3em}} $\tsigk=\mbox{std}(p)$
  \end{algorithmic}
  \end{algorithm}%
\end{minipage}%
		\begin{tikzpicture}[overlay]
		    \path[<->,thick,blue!70!black] (hopt1)
			[out=0] edge [in=180] (hopt2);
		\end{tikzpicture}%
\vspace{-1em}
\caption{
  Estimators of the performance of a method, and its variation.
We represent the seeding of sources of variations with $\xi\sim \mbox{RNG()}$, 
where RNG() is some random number generator.
Their difference lies in the hyper-parameter optimization step (HOpt).
The ideal estimator requires executing $k$ times $\Hopt$, each requiring $T$ trainings
for the hyperparameter optimization, for a total of $O(k \cdot T)$ trainings. 
The biased estimator requires executing only 1 time $\Hopt$, for $O(k + T)$ trainings in total.
\label{fig:estimators}%
}
\vspace{-1em}
\end{figure}

\subsubsection{Biased estimator: fixing HOpt}
\label{sec:biased_est}

A computationally  cheaper but biased estimator consists in re-using the hyperparameters obtained from a \emph{single} hyperparameter optimization
to generate $k$ subsequent performance measures $\hat{R}_e$ where only $\xi_O$ (or a subset of
$\xi_O$) is randomized.
This procedure is presented in Algorithm~\ref{alg:biased-estimator}.
It requires only $O(k + T)$ fittings, substantially less than the ideal estimator.
The estimator is biased with $k>1$, $\E\left[\tmuk\right]\neq\mu$.
A bias will occur when a set of hyperparameters $\lambdaopt$ are optimal for a particular
instance of $\xi_O$ but not over most others.

When we fix sources of variation $\xi$ to arbitrary values 
(e.g. random seed), 
we are conditioning the distribution of
$\hre$ on some arbitrary $\xi$.
Intuitively, holding fix some sources of variations
should reduce the variance of the whole process.
What our intuition
fails to grasp however, is that this conditioning to arbitrary $\xi$
induces a correlation between the trainings which in turns increases the variance of the estimator.
Indeed, a sum of correlated variables increases with the strength of the correlations.

Let $\Var(\hre \mid \xi)$ be the variance of the conditioned performance measures $\hre$ and
$\rho$ the average correlation among all pairs of $\hre$. The variance
of the biased estimator is then given by the following equation.
%
%
\begin{equation}
  \label{eq:tmuk-var}
  \Var(\tmuk \mid \xi) = \frac{\Var(\hre \mid \xi)}{k} +
\frac{k-1}{k}\rho\Var(\hre \mid \xi)
\end{equation}
We can see that with a large enough correlation $\rho$, the variance $\Var(\tmuk \mid \xi)$ could be dominated by
the second term. In such case, increasing the number of data splits $k$ would not reduce the
variance of $\tmuk$.
Unlike with $\hmuk$, the mean square error for $\tmuk$
will not decreases with $k$ :
\begin{align}
  \E[(\tmuk - \mu)^2] 
      &= \Var(\tmuk\mid\xi) + (\E[\tmuk \mid \xi] - \mu)^2 \nonumber \\
      &= \frac{\Var(\hre \mid \xi)}{k} + \frac{k-1}{k}\rho\Var(\hre \mid \xi) \nonumber\\
      & \qquad + (\E[\hre \mid \xi] - \mu)^2
\end{align}
This result has two implications, one beneficial to improving benchmarks,
the other not.
Bad news first: the limited effectiveness of increasing $k$ to improve the quality of the estimator
$\tmuk$ is a consequence of ignoring the variance induced by hyperparameter optimization. We
cannot avoid this loss of quality if we do not have the budget for repeated independent
hyperoptimization.
The good news is that current practices generally account for only one or 
two sources of variation; there is thus room for improvement. This
has the potential of decreasing the average correlation $\rho$ and 
moving $\tmuk$ closer to $\hmuk$. We will see empirically
in next section how accounting for more sources of variation moves us closer 
to $\hmuk$ in most of our case studies.

\subsection{The cost of ignoring $\Hopt$ variance}
\label{sec:hopt-variance}



To compare the estimators $\hmuk$ and $\tmuk$ presented above,
we measured empirically the statistics of the estimators
on budgets of $k=(1, \cdots, 100)$ points
on our five case studies. The ideal
estimator is asymptotically unbiased and therefore only one repetition
is enough to estimate $\Var(\hmuk)$ for each task. For the biased estimator
we run 20 repetitions to estimate 
$\Var(\tmuk \mid \xi)$.
We sample 20 arbitrary $\xi$ (random seeds) and compute the standard deviation of $\tmuk$ for
${k=(1,\cdots,100)}$.
We compared the biased estimator \texttt{FixedHOptEst()} while varying different subset of sources
of variations to see if randomizing more of them would help increasing the quality of the
estimator. We note \texttt{FixedHOptEst($k$,Init)} the biased estimator $\tmuk$ 
randomizing only the weights
initialization, \texttt{FixedHOptEst($k$,Data)} the biased estimator 
randomizing only data splits, and \texttt{FixedHOptEst($k$,All)} the biased estimator
randomizing all sources of variation $\xi_O$ except for 
hyperparameter optimization.


\begin{figure}[t!]
  \includegraphics[width=\columnwidth]{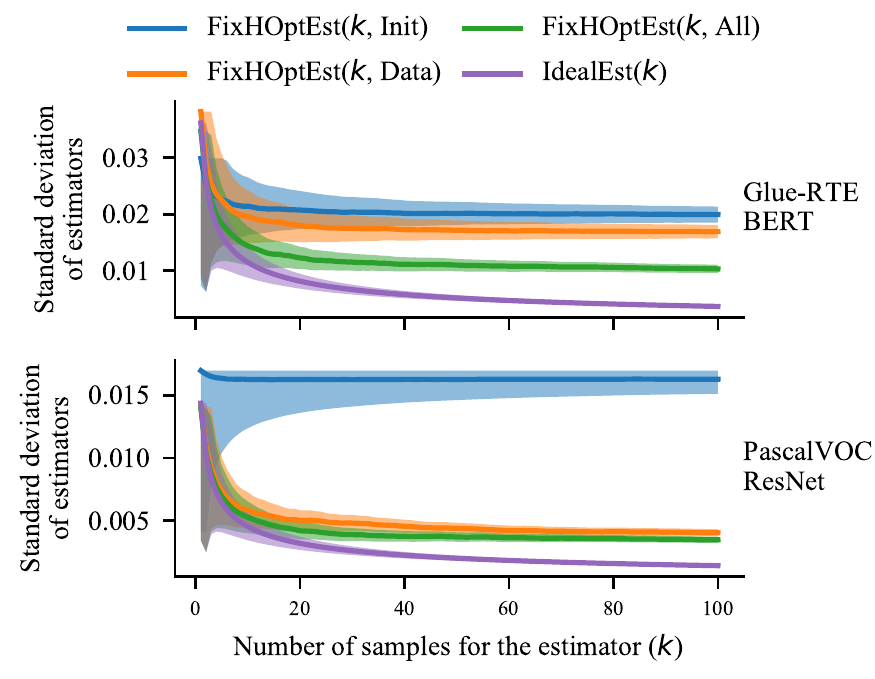}
  \vspace{-2em}
    \caption{
    \textbf{Standard error of biased and ideal estimators with $k$ samples.}
    Top figure presents results from BERT trained on RTE and bottom figure VGG11 on CIFAR10.
    All other tasks are presented in Figure~\ref{fig:standard-error-all}.
    On x axis, the number of samples used by the estimators to compute the average classification
    accuracy.
    On y axis, the standard deviation of the estimators. 
    Uncertainty represented in light color is computed analytically as the approximate standard
    deviation of the standard deviation of a normal distribution computed on $k$ samples.
    For most case studies,
    \textbf{accounting for more sources of variation reduces the standard error of $\hmuk$.}
    This is caused by the decreased correlation $\rho$ thanks to additional randomization
    in the learning pipeline.
    \texttt{FixHOptEst(k, All)} provides an improvement
    towards \texttt{IdealEst(k)} for no additional computational cost compared to
    \texttt{FixHOptEst(k, Init)} which is currently considered as a good practice.
    \textbf{Ignoring variance from $\Hopt$ is harmful for a good estimation of $\expr$.}
  }
  \label{fig:standard-error}
\end{figure}

We present results from a subset of the tasks in \autoref{fig:standard-error} (all tasks
are presented in \autoref{fig:standard-error-all}).
Randomizing weights initialization only (\texttt{FixedHOptEst($k$,init)}) provides
only a small improvement with $k>1$. In the task
where it best performs (Glue-RTE), it converges to the equivalent of $\hat{\mu}_{(k=2)}$.
This is an important result since it corresponds to the predominant approach used in the literature today.
Bootstrapping with \texttt{FixedHOptEst($k$,Data)} improves the
standard error for all tasks, converging to equivalent of $\hat{\mu}_{(k=2)}$ to $\hat{\mu}_{(k=10)}$. 
Still, the biased estimator including all sources of variations
excluding hyperparameter optimization \texttt{FixedHOptEst($k$,All)} is by far the best estimator
after the ideal estimator, converging to equivalent of $\hat{\mu}_{(k=2)}$ to $\hat{\mu}_{(k=100)}$.

This shows that accounting for all sources of variation reduces the
likelihood of error in a computationally achievable manner. \texttt{IdealEst($k=100$)} takes
1 070 hours to compute, compared to only 21 hours for each
\texttt{FixedHOptEst($k=100$)}. Our study paid the high computational
cost of multiple rounds of  \texttt{FixedHOptEst($k$,All)}, and the cost of
\texttt{IdealEst($k$)} for a total of \textit{6.4 GPU years} to show that
\texttt{FixedHOptEst($k$,All)} is better than the status-quo and a
satisfying option for statistical model comparisons \emph{without} these
prohibitive costs.




%

\section{Accounting for variance to draw reliable conclusions}
\label{sec:error-rates}

\subsection{Criteria used to conclude from benchmarks}
\label{sec:comparison-methods}


Given an estimate of the performance of two learning pipelines and their
variance, are these two pipelines different in a meaningful way?
We first formalize common practices to draw such conclusions, then
characterize their error rates.

\paragraph{Comparing the average difference}

A typical criterion to 
conclude that one algorithm is superior to another is that
one reaches a performance superior to another by some (often implicit) threshold $\delta$.
The choice of the
threshold $\delta$ can be arbitrary, but a reasonable one is to consider previous
accepted improvements, e.g. improvements in Figure~\ref{fig:paperswithcode}.

This difference in performance is sometimes computed across a single run
of the two pipelines, but a better practice used in the deep-learning
community is to average multiple seeds \cite{bouthillier:hal-02447823}.
%
Typically hyperparameter optimization is performed
for each learning algorithm and then several weights initializations or 
other sources of fluctuation are sampled, giving $k$ estimates of the
risk $\hat{R}_e$ -- note that these are biased as detailed in
\autoref{sec:biased_est}.
If an algorithm $A$ performs better than an algorithm $B$ by at least $\delta$
on average, it is considered as
a better algorithm than $B$ for the task at hand. This approach does not account for false detections and thus can not easily distinguish between true impact and random chance.  

Let $\hat{R}_e^A = \frac{1}{k}\sum_{i=1}^k \hat{R}_{ei}^{A}$,
where $\hat{R}_{ei}^{A}$ is the empirical risk of algorithm $A$ on the
$i$-th split, be the mean performance of algorithm $A$, and
similarly for $B$.
The decision whether $A$ outperforms $B$ is then determined by
$({\hat{R}_e^A - \hat{R}_e^B > \delta})$.

The variance is not accounted for in the average comparison. We will now
present a statistical test accounting for it. Both comparison methods will next
be evaluated empirically using simulations based on our case studies.
%
%
%

\paragraph{Probability of outperforming}
The choice of threshold $\delta$ 
is  problem-specific and does not relate well to a statistical
improvement. Rather, we propose to formulate the comparison in terms of
\textit{probability of improvement}.
Instead of comparing the average performances, we compare their distributions altogether.
Let $\pab$ be the probability of measuring a better performance 
for $A$ than $B$ across fluctuations such as data splits and weights initialization.
To consider an algorithm $A$ significantly better than $B$, we ask
that $A$ outperforms $B$ \emph{often enough}:
$\pab \ge \gamma$. Often enough, as
set by $\gamma$, needs to be defined by community standards, which we
will revisit below.
This probability can simply be computed as the proportion of successes, 
$\hat{R}_{ei}^A>\hat{R}_{ei}^B$, where $(\hat{R}_{ei}^A,\hat{R}_{ei}^B), i\in \{1,\ldots,k\}$
are pairs of empirical risks measured on $k$ different data splits for algorithms $A$ and $B$.
\begin{equation}
  \pab = \frac{1}{k}\sum_i^k I_{\{\hat{R}_{ei}^A>\hat{R}_{ei}^B\}}
    \label{eq:count_stats}
\end{equation}
where $I$ is the indicator function. We will build upon the non-parametric Mann-Whitney test to produce decisions about whether $\pab \geq \gamma$  \cite{perme2019confidence} .

The problem is well formulated in the
Neyman-Pearson view of statistical testing
\cite{neyman1928use,perezgonzalez2015fisher}, which requires the
explicit definition of both a null hypothesis $H_0$ to control for
\emph{statistically significant} results, and an alternative hypothesis
$H_1$ to declare results \emph{statistically meaningful}.
A \emph{statistically significant} result is one that is not explained 
by noise, the null-hypothesis $H_0: \pab=0.5$. 
With large enough sample size, any arbitrarily small difference
can be made \emph{statistically significant}.
A \textit{statistically meaningful} result is one large enough 
to satisfy the alternative hypothesis $H_1:\pab=\gamma$.
Recall that $\gamma$ is a threshold that needs to be defined by community standards.
We will discuss reasonable values for $\gamma$ in next section based on our
simulations.

We recommend to conclude that algorithm $A$ is better than $B$ on a given task if 
the result is both \emph{statistically significant}
and \textit{meaningful}. 
The reliability of the estimation of $\pab$ can be quantified
using confidence intervals, computed
with the non-parametric percentile bootstrap \cite{efron1982jackknife}.
The lower bound of 
the confidence interval $\mbox{CI}_{\min}$
controls if the result is \emph{significant} $(\mathbb{P}(A>B) - \mbox{CI}_{\min} > 0.5)$,
and the upper bound of the confidence interval $\mbox{CI}_{\max}$ 
controls if the result is \emph{meaningful} $(\mathbb{P}(A>B) + \mbox{CI}_{\max} > \gamma)$.



\subsection{Characterizing errors of these conclusion criteria}
\label{sec:error_rates}

We now run an empirical study of the two conclusion criteria presented above, the popular
\emph{comparison of average differences} and our recommended \emph{probability of outperforming}.
We will re-use mean and variance estimates from \autoref{sec:hopt-variance} with the ideal and biased
estimators to simulate performances of trained algorithms
so that we can measure the reliability of these conclusion criteria when using ideal or biased estimators.

\paragraph{Simulation of algorithm performances}

We simulate realizations of the ideal estimator $\hmuk$ and the biased estimator
$\tmuk$ with a budget of $k=50$ data splits.
For the ideal estimator, we model $\hmuk$ with a normal distribution 
$\hmuk \sim \mathcal{N}(\mu, \frac{\sigma^2}{k})$, where $\sigma^2$
is the variance measured with the ideal estimator in our case studies, and 
$\mu$ is the empirical risk $\hre$. Our experiments consist in varying
the difference in $\mu$ for the two algorithms, to 
span from identical to widely different performance ($\mu_A >> \mu_B$).

For the biased estimator, we rely on a two stage sampling process for the simulation.
First, we sample the bias of $\tmuk$ based on the variance $\Var(\tmuk \mid \xi)$ 
measured in our case studies,
$Bias \sim \mathcal{N}(0, \Var(\tmuk \mid \xi))$. Given $b$, a sample of $Bias$,
we sample $k$ empirical risks following 
$\hre \sim \mathcal{N}(\mu + b, \Var(\hre \mid \xi))$, where 
$\Var(\hre \mid \xi)$ is the variance of the empirical risk $\hre$ averaged across
20 realizations of $\tmuk$ that we measured in our case studies. 

In simulation we vary the mean performance of $A$ with respect to the mean performance of $B$ so that
$\pab$ varies from 0.4 to 1 to test three regions:
\begin{description}
  \item[$H_0$ is true]: Not significant, not meaningful\hspace*{\fill}\\
     ${\mathbb{P}(A>B) - \mbox{CI}_{\min} \leq 0.5}$
   \item[$H_0$ \& $H_1$ are false ($\cancel{H_0} \cancel{H_1}$)]: Significant, not meaningful\hspace*{\fill}\\
    ${\mathbb{P}(A>B) - \mbox{CI}_{\min} > 0.5 \wedge \mathbb{P}(A>B) + \mbox{CI}_{\min} \leq \gamma}$
  \item[$H_1$ is true]: Significant \textit{and} meaningful\hspace*{\fill}\\
    ${\mathbb{P}(A>B) - \mbox{CI}_{\min} > 0.5 \wedge \mathbb{P}(A>B) + \mbox{CI}_{\min} > \gamma}$
\end{description}

For decisions based on comparing averages, we set $\delta=1.9952\sigma$
where $\sigma$ is the standard deviation measured in our case
studies with the ideal estimator. The value 1.9952 is set by linear regression so that 
$\delta$ matches the average improvements obtained from \url{paperswithcode.com}.
This provides a threshold $\delta$ representative of the published improvements.
For the probability of outperforming, we use a threshold of $\gamma=0.75$ which we have observed 
to be robust across all case studies (See Appendix~\ref{sec:test-analysis}).

\paragraph{Observations}

\begin{figure}[t]
  \includegraphics[width=\columnwidth]{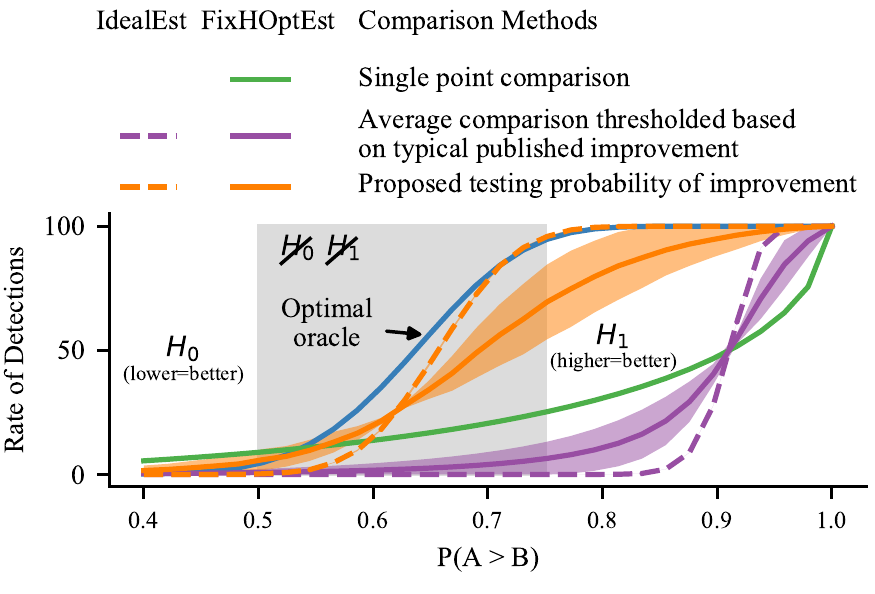}
  \vspace{-2em}
    \caption{
    \textbf{Rate of detections of different comparison methods.}
    x axis is the true simulated probability of a learning algorithm $A$ to outperform another algorithm
    $B$ across random fluctuations (ex: random data splits).
    We vary the mean performance of $A$ with respect to that of $B$ so that
    $\pab$ varies from 0.4 to 1. The blue line is the optimal oracle,
    with perfect knowledge of the variances.
    The single-point comparison (green line) has both a high rate of false positives
    in the left region ($\approx10\%$) and a high rate of false negative on the right 
    ($\approx75\%$).
    The orange and purple lines show the results for the \textit{average
comparison method} (prevalent in the literature) and our proposed \textit{probability of outperforming} method respectively. The solid versions are using the expensive ideal estimator, and the dashed line our $51\times$ cheaper, but biased, estimator. 
    The average comparison is highly conservative with a low rate of false positives ($<5\%$) on the left
    and a high rate of false negative on the right
    ($\approx90\%$), even with the expensive and exhaustive simulation.
    Using the probability of outperforming has both a reasonable rate of false positives ($\approx 5\%$) on 
    the left and a reasonable rate of false negatives on the right 
    ($\approx 30\%$) even when using our biased estimator, and approaches the oracle when using the expensive estimator.
    }
  \label{fig:error-rates}
\end{figure}

Figure~\ref{fig:error-rates}
reports results for different decision criteria, using
the ideal estimator and the biased estimator, as the difference in
performance of the algorithms $A$ and $B$ increases (x-axis). 
The x-axis is broken into three regions: 1) Leftmost is when $H_0$ is true (not-significant). 2) The grey
middle when the result is significant, but not meaningful in our framework  ($\cancel{H_0} \cancel{H_1}$). 3) The rightmost is
when $H_1$ is true (significant and meaningful). 
The single point comparison leads to the worst decision by far.
It suffers from both high false positives ($\approx10\%$)
and high false negatives ($\approx75\%$). The average with $k=50$,
on the other hand,
is very conservative with low false positives ($<5\%$) but very high false negatives 
($\approx90\%$). Using the probability of outperforming leads to better
balanced decisions, with 
a reasonable rate of false positives ($\approx 5\%$) on the left and a reasonable rate of false
negatives on the right ($\approx 30\%$).

The main problem with the average comparison is the threshold. 
A t-test 
only differs from an average in that the threshold
is computed based on the variance of the model performances and the sample size.
It is this adjustment of the threshold based on the variance that allows better control on false
negatives.

Finally, we observe that the test of probability of outperforming ($\pab$) controls
well the error rates even when used with a biased estimator.
Its performance is nevertheless 
impacted by the biased
estimator compared to the ideal estimator. Although we cannot guarantee a
nominal control, we confirm that it is a major improvement compared to the commonly
used comparison method at no additional cost. 

\section{Our recommendations: good benchmarks with a budget}
\label{sec:recommendations}

We now distill from the theoretical and empirical results of the previous
sections a set of practical recommendations to benchmark
machine-learning pipelines. Our recommendations are pragmatic in the
sense that they are simple to implement and cater for limited
computational budgets. 

\paragraph{Randomize as many sources of variations as possible}
Fitting and evaluating a modern machine-learning pipeline comes with many
arbitrary aspects, such as the choice of initializations or the data
order. Benchmarking a pipeline given a specific instance of these choices
will not give an evaluation that generalize to new data, even drawn from
the same distribution. On the opposite, a benchmark that varies these
arbitrary choices will not only evaluate the associated variance 
(section \ref{sec:variance}), but also reduce the error on the expected
performance as they enable measures of performance on the test set that are
less correlated (\ref{sec:estimation}). This counter-intuitive phenomenon
is related to the variance reduction of bagging
\cite{breiman1996bagging,buhlmann2002analyzing}, and helps characterizing
better the expected behavior of a machine-learning pipeline, as opposed
to a specific fit.


\paragraph{Use multiple data splits}
The subset of the data used as test set to validate an algorithm is
arbitrary. As it is of a limited size, it comes with a limited estimation
quality with regards to the performance of the algorithm on wider samples
of the same data distribution (figure \ref{fig:binomial_model}).
Improvements smaller than this variance observed on a given test set will
not generalize. Importantly, this variance is not negligible compared to
typical published improvements or other sources of variance
(figures \ref{fig:variances} and \ref{fig:paperswithcode}).
For pipeline comparisons with more statistical power, it is useful to draw multiple tests, for
instance generating random splits with a out-of-bootstrap scheme
(detailed in
appendix \ref{sec:bootstrap-appendix}). 

\paragraph{Account for variance to detect meaningful improvements}
Concluding on the significance --statistical or practical-- of an
improvement based on the difference between average performance
requires the choice of a threshold that can be difficult to set.
A natural scale for the threshold is the variance of the benchmark, but this
variance is often unknown before running the experiments.
Using the probability of outperforming $\pab$ with a 
threshold of $0.75$ gives empirically a criterion
that separates well benchmarking fluctuations from published improvements
over the 5 case studies that we considered. We recommend to always
highlight not only the best-performing procedure, but also all those
within the significance bounds.
We provide an example in Appendix~\ref{sec:statistical-test-appendix}
to illustrate the application of our recommended statistical test.

\section{Additional considerations}
\label{sec:additional_considerations}

There are many aspects of benchmarks which our study has not addressed.
For completeness, we discuss them here.

\paragraph{Comparing models instead of procedures} Our framework provides value when the user can control the model training process and source of variation. In cases where models are \textit{given} but not under our control (e.g., purchased via API or a competition), 
the only source of variation left is the data used to test the model. Our framework 
and analysis does not apply to such scenarios.

\paragraph{Benchmarks and competitions with many contestants}

We focused on comparing two learning algorithms. 
Benchmarks -- and competitions in particular -- commonly involve large number of learning
algorithms that are being compared. Part of our results carry over
unchanged in such settings, in particular those related to variance and
performance estimation. With regards to reaching a well-controlled
decision, a new challenge comes from multiple comparisons when there are
many algorithms. A possible alley would be to adjust the decision
threshold $\gamma$, raising it with a correction for multiple comparisons
(e.g. Bonferroni)
\cite{dudoit2003multiple}. However, as the number gets larger, the
correction becomes stringent. In competitions where the number of
contestants can reach hundreds, the choice of a winner comes necessarily
with some arbitrariness: a different choice of test sets might have led
to a slightly modified ranking.

\paragraph{Comparisons across multiple dataset }
Comparison over multiple datasets is often used to accumulate evidence
that one algorithm outperforms another one. The challenge is to
account for different errors, in particular different levels of variance,
on each dataset.

\citet{demvsar2006statistical} recommended Wilcoxon signed
ranks test or Friedman tests to compare classifiers across multiple datasets. 
These recommendations are however hardly applicable on small sets of datasets -- 
machine learning works typically include as few as 3 to 5 datasets \cite{bouthillier:hal-02447823}.
The number of datasets corresponds to the sample size of these tests, and such a small
sample size leads to tests of very limited statistical power.

\citet{dror2017replicability} propose to accept methods that give
improvements on \emph{all} datasets, controlling for multiple
comparisons. As opposed to \citet{demvsar2006statistical}'s
recommendation, this approach performs well with a small number of datasets.
On the other hand, a large number of datasets
will increase significantly the severity of the family-wise error-rate correction, making
Dem\v{s}ar's recommendations more favorable.

\paragraph{Non-normal metrics}

We focused on model performance, but model evaluation in practice 
can include other metrics such as the training time to reach a performance level 
or the memory foot-print \cite{reddi2020mlperf}.
Performance metrics are generally averages over samples 
which typically makes them amenable to a reasonable normality assumption.

\section{Conclusion}

We showed that fluctuations in the performance measured by
machine-learning benchmarks arise from many different sources. In deep
learning, most evaluations focus on the effect of random
weight initialization, which actually contribute a
small part of the variance, on par with residual fluctuations of 
hyperparameter choices after their optimization but much smaller than
the variance due to perturbing the split of the data in train and test
sets.
Our study clearly shows that these factors must be accounted to give
reliable benchmarks. For this purpose, we study estimators of benchmark variance
as well as decision criterion to conclude on an improvement. 
Our findings outline recommendations to improve reliability of
machine learning benchmarks:
1) randomize as many sources of variations as possible in the performance
estimation;
2) prefer multiple random splits to fixed test sets;
3) account for the resulting variance when concluding on the benefit of
an algorithm over another.

\nocite{HutHooLey14}

\bibliography{main}
\bibliographystyle{mlsys2020}

\appendix

\section{Notes on reproducibility}
\label{sec:repro-appendix}

Ensuring full reproducibility is often a tedious work. We provide here notes and remarks
on the issues we encountered while working towards fully reproducible experiments.

\paragraph{The testing procedure}
To ensure proper study of the sources of variation it was necessary to control them
close to perfection. For all tasks, we ran a pipeline of tests to ensure perfect 
reproducibility at execution and also at resumption. During the tests, 
each source of variation was varied with 5 different seeds, each executed 5 times.
This ensured that the pipeline was reproducible for different seeds. Additionally,
for each source of variation and for each seed,
another training was executed but 
automatically interrupted after each epoch. The worker would then start the training of the next seed 
and iterate through the trainings for all seeds before resuming the first one. All these
tests uncovered many bugs and typical reproducibility issues in machine learning.
We report here some notes.

\paragraph{Computer architecture \& drivers}
Although we did not measure the variance induced by different GPU architectures,
we did observe that different GPU models would lead to different results.
The CPU model had less impact on the Deep Learning tasks but the MLP-MHC task
was sensitive to it. We therefore limited all tasks to specific computer architectures.
We also observed issues when CUDA drivers were updated during preliminary 
experiments. We ensured all experiments were run using CUDA 10.2.

\paragraph{Software \& seeds}

PyTorch versions lead to different results as well. We ran every
Deep Learning experiments with PyTorch 1.2.0.

We implemented our data pipeline so that we could seed the iterators,
the data augmentation objects and the splitting of the datasets.
We had less control at the level of the models however. For PyTorch 1.2.0,
the random number generator (RNG) must be seeded globally which 
makes it difficult to seed different parts separately. We seeded PyTorch's global RNG for
weight initialization at the beginning of the training process and then seeded PyTorch's RNG
for the dropout. Afterwards we checkpoint the RNG state so that we can restore the RNG states
at resumption. We found that models with convolutionnal layers would not yield reproducible
results unless we enabled \texttt{cudnn.deterministic} and disabled
\texttt{cudnn.benchmark}.

We used the library RoBO \cite{klein-bayesopt17} for our Bayesian Optimizer. There was
no support for seeding, we therefore resorted to seeding the global seed of python and numpy random
number generators.
We needed again to keep track of the RNG states and checkpoint them so that we can resume
the Bayesian Optimizer without harming the reproducibility.

For one of our case study, image segmentation, we have been unable to make
the learning pipeline perfectly reproducible. This is problematic because
it prevents us from studying each source of variation in isolation. We thus
trained our model with every seeds fixed across all 200 trainings and measured the variance
we could not control. This is represented as the numerical noise in Figures~\ref{fig:variances}
and \ref{fig:normal}.
%
%



\todo{Complete}


\setcounter{section}{1}

\renewcommand{\thefigure}{\thesection.\arabic{figure}}
\setcounter{figure}{0}

\section{Our bootstrap procedure}
\label{sec:bootstrap-appendix}

Cross-validation with different $k$ impacts the number of samples, it is not the case
with not bootstrap.
That means flexible sample sizes for statistical tests is hardly possible with
cross-validation within affecting the training dataset sizes. 
\cite{hothorn2005design} focuses on the dataset sampling as the most important
source of variation and marginalize out all other sources by taking the average
performance over multiple runs for a given dataset. This increases even more
the computational cost of the statistical tests.

We probe the effect of data sampling with bootstrap, specifically by
bootstrapping to generate training sets and measuring the
out-of-bootstrap error, as introduced by \citet{breiman1996out} in the
context of bagging and generalized by \cite{hothorn2005design}. For completeness,
we formalize this use of the bootstrap to create training and test sets
and how it can estimate the variance of performance measure
due to data sampling on a finite dataset. 



We assume we are seeking to generate sets of i.i.d. samples from true distribution $\dist$.
Ideally we would have access to $\dist$ and could sample our
finite datasets independently from it.
\begin{equation}
  S^t_b = \{(x_1, y_1), (x_2, y_2), \cdots (x_n, y_n)\} \sim \dist
\end{equation}
Instead we have one dataset $S \sim \dist^n$ of finite size $n$ and need to sample independent
datasets from it. A popular method in machine learning to estimate performance on a small
dataset is cross-validation
\cite{bouckaert2004evaluating, dietterich1998approximate}. This
method however underestimates variance because of correlations induced by the process.
We instead favor bootstrapping \cite{efron1979} as used by \cite{hothorn2005design} to 
simulate independent data sampling from the true distribution.
\begin{equation}
  S^t_b = \{(x_1, y_1), (x_2, y_2), \cdots (x_n, y_n)\} \sim S
\end{equation}
Where $S^t_b \sim S$ represents sampling the $b$-th
training set with replacement from the set $S$.
We then turn to out-of-bootstrapping to generate the held-out set. We use 
all remaining samples in $S \setminus S^t_b$ to sample $S^o_b$.
\begin{equation}
  S^o_b = \{(x_1, y_1), (x_2, y_2), \cdots (x_n, y_n)\} \sim S \setminus S^t_b
\end{equation}
This procedure is represented as $(S^{tv}, S^o)\sim sp_{n,n^{\prime}}(S)$ in 
the empirical average risk $\hat{R}_\mathcal{P}(S, n,n')$, end of 
Section~\ref{sec:learning_framework}.

\section{Statistical testing}
\label{sec:statistical-test-appendix}

We are interested in asserting whether a learning algorithm $A$ better performs than 
another learning algorithm $B$. Measuring the performance of these learning algorithms
is not a deterministic process however and we may be deceived if noise is not accounted for.
Because of the noise, we cannot know for sure whether a conclusion we draw is true,
but using a statistical test, we can at least ensure a bounded rate of false positives
(drawing $A>B$ while truth is $A\le B$) and false negatives 
(drawing $A\le B$ while truth is $A > B$).
The capacity of a statistical test to identify true differences, that is, of correctly
inferring $A>B$ when this is true, is called the statistical power of a test.
%
%
The procedure we describe here seeks to avoid deception from false positives while 
providing a strong statistical power.

We will describe the entire procedure, from the generation of the performance measures 
(Sections~\ref{sec:stat-randomize-appendix} \& \ref{sec:stat-pairing-appendix}),
the estimation of sample size (Section~\ref{sec:stat-sample-size-appendix}),
computation of $\pab$ (Section~\ref{sec:stat-pab-appendix}), computation of the
confidence interval (Section~\ref{sec:stat-ci-appendix}) to the inference
based on the statistical test (Section~\ref{sec:stat-conclusion-appendix})

\subsection{Randomizing sources of variance}
\label{sec:stat-randomize-appendix}

As shown in Section~\ref{sec:estimation}, randomizing as many sources of variance as possible in the
learning pipelines help reduce the correlation and thus improve the reliability of the performance
estimation. The simplest way to randomize as many as possible is to simply avoid seeding the random 
number generators. We list here sources of variations we faced in our case studies,
but there exists many other sources of variations in diverse learning algorithms and tasks.

\begin{description}
  \item[Data splits]
    The data being used should ideally always be different samples from the true
    distribution of interest. In practice we only have access to a finite dataset and 
    therefore the best we can do is random splits with cross-validation or out-of-bootstrap
    as described in Appendix~\ref{sec:bootstrap-appendix}.
  \item[Data order]
    The ordering of the data can have a surprisingly important impact as can be
    observed in Figure~\ref{fig:variances}.
  \item[Data augmentation]
    Stochastic data augmentation should not be seeded, so that it follows a different sequence 
    at each run.
  \item[Model initialization]
    Model initialization, e.g. weights initialization in neural networks, should be randomized
    across all trainings.
  \item[Model stochasticity]
    Learning algorithms sometimes include stochastic computations such as dropout in neural
    networks \cite{JMLR:v15:srivastava14a}, or samplings 
    methods \cite{DBLP:journals/corr/KingmaW13, DBLP:conf/iclr/MaddisonMT17}.
  \item[Hyperparameter optimization]
    The optimization of the hyperparameters generally include stochasticity which should ideally
    be randomized. Running multiple hyperparameter optimizations may often be practically
    unaffordable. Tests may still be carried out while fixing the hyperparameters after
    a single hyperparameter optimization, but keep in mind the incurred degradation of 
    the reliability of the conclusion as shown in Section~\ref{sec:error-rates}.
\end{description}

\subsection{Pairing}
\label{sec:stat-pairing-appendix}

Pairing is optional but is highly recommended to increase statistical power.
Avoiding seeding is the simplest solution for the randomization, but it is not the best solution. 
If possible, meticulously seeding all sources of variation with different random seeds at each 
run makes it possible to pair trainings of the algorithms so that we can conduct paired 
comparisons.

Pairing is a simple but powerful way of increasing the power of statistical tests, that is,
enabling
the reliable detection of difference with smaller sample sizes. Let $\sigma_A$ and $\sigma_B$ 
be the
standard deviation of the performance metric of learning algorithms $A$ and $B$ respectively.
If measures of $\hat{R}_e^A$ and $\hat{R}_e^A$ are not paired, the standard deviation of
$\hat{R}_e^A - \hat{R}_e^B$ is then $\sigma_A + \sigma_B$. If we pair them, then 
we marginalize out sources of
variance which results in a smaller variance $\sigma_{A-B} \leq \sigma_A + \sigma_B$. This reduction of
variance makes it possible to reliably detect smaller differences without increasing the sample
size. 


To pair the learning algorithms, sources of variation should be randomized similarly for all of
them. For instance, the random split of the dataset obtained from out-of-bootstrap should be used
for both $A$ and $B$ when making a comparison. Suppose we plan to execute 10 runs of $A$ and $B$,
then we should generate 10 different splits
$\{(S_1^{tv}, S_1^o), (S_2^{tv}, S_2^o), \cdots, (S_{10}^{tv}, S_{10}^o) \}$
and train $A$ and $B$ on each. The performances $(\hat{R}_{ei}^A, \hat{R}_{ei}^B)$
would then be compared only on the corresponding splits $(S_i^{tv}, S_i^o)$.
The same would apply to all other sources of variations. In practical terms, pairing
$A$ and $B$ requires sampling seeds for each pairs,
re-using the same seed for $A$ and $B$ in each pairs.

For some sources of variation it may not make sense to pair. This is the case
for instance with weights initialization
if $A$ and $B$ involve different neural network architectures. We can still
pair. This would not help much, but would not hurt as well. In doubt, it is better to pair.

\subsection{Sample size}
\label{sec:stat-sample-size-appendix}

As explained in Section~\ref{sec:estimation}, the more runs we have from 
$\hat{R}_e^A$ and $\hat{R}_e^B$, the more reliable the estimate of $\pab$ is.
Lets note this number of runs as the sample size $N$, not to be confused with dataset size $n$.
There exist a way of computing the minimal sample size required to ensure a minimal rate of 
false negatives based on power analysis.


We must first set the threshold $\gamma$ for our test. Based on our experiments
in Section~\ref{sec:error-rates}, we recommend a value of 0.75. We then set the desired rates
of false positives and false negatives with $\alpha$ and $\beta$ respectively.
Usual value for $\alpha$ is  0.05 while $\beta$ ranges from
0.05 to 0.2. We recommend $\beta=0.05$ for a strong statistical power.

The estimation of $P(A>B)$ is equivalent to a Mann–Whitney
test~\cite{perme2019confidence}, thus we can use Noether's sample size determination method
for this type of test \cite{noether1987sample}.

\begin{equation*}
  \label{eq:sample_size-appendix}
  N \geq  \left(\frac{\Phi^{-1}(1-\alpha) - \Phi^{-1}(\beta)}{\sqrt{6}(\frac{1}{2} - \gamma)}\right)^2
\end{equation*}

Where $\Phi^{-1}$ is the inverse cumulative function of the normal distribution.

\begin{figure}
  \center
  \includegraphics[width=\columnwidth]{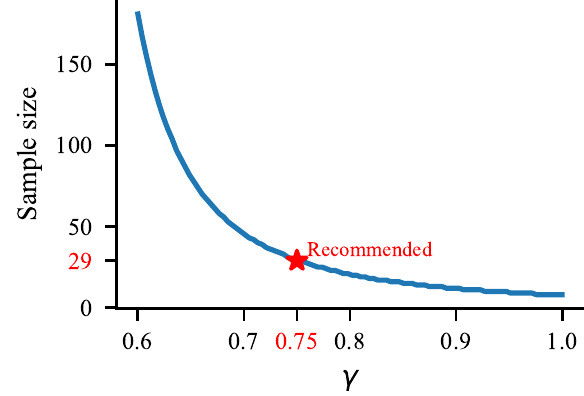}
    \caption{
    \textbf{Minimum sample size to detect ${P(A>B)>\gamma}$ reliably.}
     x-axis is the threshold $\gamma$ and y-axis is the minimum sample size 
     to reliably detect ${P(A>B)>\gamma}$. The red star shows the recommended threshold $\gamma$
     based on our results in Section~\ref{sec:error-rates} and the corresponding
     minimal sample size. We see that detecting reliably ${P(A>B)<0.6}$ is unpractical
     with minimal sample sizes quickly moving above 500. The recommended threshold on the
     other hand leads to a reasonable sample size of 29.
  }
  \label{fig:sample-size}
\end{figure}

Figure~\ref{fig:sample-size} shows how the minimal sample size evolves with $\gamma$.
Detecting $P(A>B)$ below $\gamma=0.6$ is unpractical, requiring more that 700 trainings
below 0.55 for instance. For a threshold that is representative of the 
published improvements as presented
in Figure~\ref{fig:paperswithcode}, $\gamma=0.75$, the minimal sample size required
to ensure a rate of 5\% false negatives (as defined by $\beta=0.05$) is reasonably small; 29 trainings.

\subsection{Compute $\pab$}
\label{sec:stat-pab-appendix}

For all paired performances $(\hat{R}_{ei}^A, \hat{R}_{ei}^B)$, we 
compute $I_{\{\hat{R}_{ei}^A, \hat{R}_{ei}^B\}}$, where $I$ is the indicator function.
If trainings were not paired as described in Section~\ref{sec:stat-pairing-appendix},
the pairs are randomly selected. We can then compute $\pab$ following
Equation~\ref{eq:count_stats}.

\subsection{Confidence interval of $\pab$ with percentile bootstrap}
\label{sec:stat-ci-appendix}

For the estimation of $\pab$ with values below 0.95, we recommend the use of the
the percentile bootstrap\footnote{
Percentile bootstrap is not always reliable depending on the underlying distribution 
and resampling methods but should generally be good for distributions of $\pab$ of
learning algorithms below 0.95. See \cite{canty2006bootstrap} for a discussion on the topic.}
\cite{efron1994introduction}.

Suppose we have $N$ pairs $(\hat{R}_{ei}^A, \hat{R}_{ei}^B)$. 
To compute the percentile bootstrap, we first generate $K$ groups of $N$ pairs.
To do so, we sample with replacement $N$ pairs, and do so independently $K$ times.
For each of the $K$ groups, we compute $\pab$. We sort the $K$ estimations of $\pab$ and
pick the $\alpha/2$-percentile
and $(1 - \alpha/2)$- percentile as the lower and upper bounds. The confidence interval
is defined as these lower and upper bounds computed with percentile bootstrap.

\subsection{Statistical test with $\pab$}
\label{sec:stat-conclusion-appendix}

Let $\mbox{CI}_{\min}$ and $\mbox{CI}_{\max}$ be the lower and upper bounds of the confidence
interval. We draw a conclusion based on the three following scenarios.

\begin{description}
  \item[$\mbox{CI}_{\min} \le 0.5$]: Not statistically significant. No conclusion should be drawn as
    the result could be explained by noise alone.
  \item[$\mbox{CI}_{\max} \le \gamma$]: Not statistically meaningful. Perhaps $\mbox{CI}_{\min}>
    0.5$ but it is irrelevant since $\pab$ is too small to be meaningful.
  \item[$\mbox{CI}_{\min} > 0.5  \wedge \mbox{CI}_{\max} > \gamma$]: Statistically significant and
    meaningful. We can conclude that learning algorithm $A$ is better performing
    than $B$ in the conditions defined by the experiments.
\end{description}

\section{Case studies}
\label{sec:case-studies-appendix}

\subsection{CIFAR10 Image classification with VGG11}

\paragraph{Task} CIFAR10 \cite{krizhevsky2009learning} is a dataset of 60,000 32x32 color images 
selected from 80 million tiny images dataset \cite{torralba200880}, 
divided in 10 balanced classes. The original split contains 50,000 images for training 
and 10,000 images for testing. We applied random cropping and random horizontal flipping
data augmentations.

\paragraph{Bootstrapping} The aggregation of all original 
training and testing samples are used for the bootstrap.
To preserve the balance of the classes, we applied stratified
bootstrap. For each class separately, we sampled with replacement 4,000 training samples, 
1,000 for validation and 1,000 for testing. As for all tasks, we use out-of-bootstrap
to ensure samples cannot be contained in more than one set. 

\paragraph{Model} We used VGG11 \cite{simonyan2014very} with batch-normalization and no dropout.
The weights are initialized with Glorot method based on a uniform distribution 
\cite{pmlr-v9-glorot10a}.

\paragraph{Search space for hyperparameters}

We focused on learning rate, weight decay, momentum and learning rate schedule. Batch-size was
omitted to simplify the multi-model training on GPUs, so that memory usage was consistent and
predictable across all hyperparameter settings. To ease the definition of the search space for the
learning rate schedule, we used exponential decay instead of multi-step decay despite
the wide use of the latter with similar tasks and models 
\cite{simonyan2014very,xie2019self,mahajan2018exploring,Liu_2018_ECCV,he2016identity,he2015delving}.
The former only require tuning of $\gamma$ while the later requires additionally selecting number of steps.
Search space for all experiments and default values used for 
the variance experiments are presented in
Table~\ref{table:hps-space-appendix}.



\begin{table}[t]
\centering
\caption{Computational infrastructure for CIFAR10-VGG11 experiments.}
\label{table:comp-architechture-appendix}
\begin{tabular}{lll}
Hardware/Software  & Type/Version \\\hline
CPU     & Intel(R) Xeon(R) Gold \\
        & 6148 CPU @ 2.40GHz \\
GPU model & Tesla V100-SXM2-16GB\\
GPU driver & 440.33.01 \\
OS      & CentOS 7.7.1908 Core \\
Python  & 3.6.3 \\
PyTorch &  1.2.0 \\
CUDA    &  10.2
\end{tabular}
\end{table}


\begin{table}[t]
\centering
\caption{Search space and default values for the hyperparameters in CIFAR10-VGG11 experiments.}
\begin{tabular}{lll}
  Hyperparameters   & Default & Space  \\\hline
  learning rate      & 0.03    & log($0.001$, $0.3$) \\
  weight decay       & 0.002   & log($10^{-6}$, $10^{-2}$) \\
  momentum           & 0.9     & lin($0.5$, $0.99$) \\
  $\gamma$ of lr schedule & 0.97  & lin($0.96$, $0.999$) \\
  batch-size & 128 & -
\end{tabular}
\label{table:hps-space-appendix}
\end{table}

\begin{table}[t]
\centering
\caption{Search space and default values for the hyperparameters in SST-2/RTE-BERT experiments. }
\begin{tabular}{lll}
  Hyperparameters       & Default & Space  \\\hline
  learning rate         & $2*10^{-5}$ & log($10^{-5}$, $10^{-4}$) \\
  weight decay          & $0.0$ & log($10^{-4}$, $2*10^{-3}$) \\
  std for weights init. & $0.2$ & log($0.01$, $0.5$) \\
  $\beta_1$ &  0.9 & - \\
  $\beta_2$ &  0.999 & - \\
  dropout rate & $0.1$ & - \\
  batch size & 32 & -
\end{tabular}
\label{table:hps-sst2-rte-space-appendix}
\end{table}

\subsection{Glue-SST2 sentiment prediction with BERT}

\paragraph{Task} SST2 (Stanford Sentiment Treebank) \cite{socher2013recursive} is a binary
classification task included in GLUE \cite{wang2019glue}. In this task, the input is a sentence from
a collection of movie reviews, and the target is the associated sentiment (either positive or
negative). The publicly available data contains around 68k entries.

\paragraph{Bootstrapping} We maintained the same size
ratio between train/validation (i.e., 0.013) when performing the bootstrapping analysis. 
We performed standard out-of-bootstrap without conserving class balance since
the original dataset is not balanced and ratios between classes vary from training and validation
set in the original splits. The variable ratios of classes across bootstrap samples 
generate additional variance in our results, but is representative of the effect of generating a
dataset that is not perfectly balanced.

\paragraph{Model} We used the BERT \cite{devlin2018bert} implementation provided by the Hugging Face
\cite{Wolf2019HuggingFacesTS} repository. BERT is a Transformer \cite{vaswani2017attention} encoder
pre-trained on the self-supervised Masked Language Model task \cite{devlin2018bert}.  We chose BERT
given its importance and influence in the NLP literature.  It is worthy to note that the
pre-training phase of BERT is also affected by sources of variations. Nevertheless, we didn't
investigate this phase given the amount of time (and resources) required to perform it. Instead, we
always start from the (same) pre-trained model image provided by the Hugging Face
\cite{Wolf2019HuggingFacesTS} repository. Indeed, the weight initialization was only applied to the final classifier. The initialization method used is standard Gaussian with $0.0$ mean and standard deviation that depends on the related hyperparameter.

\paragraph{Search space of hyperparameters}
We ran a small-scale hyperparameter space exploration in order to select the hyperparameter
search space to use in our experiments.
As such, we decided to include the learning rate, weight decay and the standard deviation for the
model parameter initialization (see Table~\ref{table:hps-sst2-rte-space-appendix}).
We fixed the dropout probability to the value of 0.1 as in the original BERT architecture.
For the same reason, we fixed $\beta_1=0.9$ and $\beta_2=0.999$.
Default values used for the variance experiments are also reported in
Table~\ref{table:hps-sst2-rte-space-appendix}.
The model has been fine-tuned on SST2 for 3 epochs, with a batch size of 32. Training has been performed with mixed precision.
Note that for weight decay we used the default value from the Hugging Face repository (i.e., $0.0$)
even if this is outside of the hyperparameter search space. We confirmed that this makes no
difference by looking at the results of the small-scale hyperparameter space exploration.

\subsection{Glue-RTE entailment prediction with BERT}

\paragraph{Task} RTE (Recognizing Textual Entailment) \cite{bentivogli2009fifth} is a also a binary
classification task included in GLUE \cite{wang2019glue}. The task is a collection of text fragment
pairs, and the target is to predict if the first text fragment entails the second one. RTE dataset
only contains around 2.5k entries.

\paragraph{Bootstrapping} In our bootstrapping analysis we maintained the train/validation
ratio of 0.1. As for Glue-SST2, we used standard out-of-bootstrap and did not preserve original class
ratios.

\paragraph{Model \& search space of hyperparameters}
We used the BERT \cite{devlin2018bert} model
for RTE as well, trained in the same way specified in the SST-2 section. In particular, we used the
same hyperparameters (see Table \ref{table:hps-sst2-rte-space-appendix}), same batch size, and we
trained in the same mixed-precision environment. The model has been fine-tuned on RTE for 3 epochs.

\subsection{PascalVOC image segmentation with ResNet Backbone}

\paragraph{Task} The PascalVOC segmentation task \cite{pascal-voc-2012} entails generating
pixel-level segmentations to classify each pixel in an image as one of 20 classes or background.
This publicly available dataset contains 2913 images and associated ground truth segmentation
labels. The original splits contains 2184 images for training and 729 for validation.
Images were
normalized and zero-padded to a final size of 512x512.

\paragraph{Bootstrapping} We used a train/validation ratio of 0.25 for our bootstrap analysis,
generating training sets of 2184 images, validation and test sets of 729 images each. Since multiple
classes can appear in a single image, the original dataset was not balanced, we thus used standard
out-of-bootstrap for our experiments.

\paragraph{Model} We used an FCN-16s \cite{long2014fully} with a ResNet18 backbone \cite{he2015deep}
pretrained on ImageNet \cite{imagenet_cvpr09}. After exploring several possible backbones, ResNet18
was selected since it could be trained relatively quickly.
We use weighted cross entropy, with only predictions within the original image boundary contributing to the loss. The model is optimized using SGD with momentum.

\paragraph{Metric} The metric used is the mean Intersection over Union (mIoU) of the twenty classes and the background class. The complement of the mIoU, the mean Jaccard Distance, is the metric minimized in all HPO experiments. 

\paragraph{Search space of hyperparameters} Certain hyperparmeters, such as the number of kernals,
or the total number of layers, are part of the definition of the ResNet18 architecture.
As a result, we explored key optimization hyperparameters including: learning rate, momentum,
and weight decay. The hyperparameter ranges selected, as well as the default hyperparameters used in
the variance experiments, can be found in table \ref{table:hps-pascal-voc-space-appendix} and in
table \ref{table:hps-pascal-voc-default-appendix}, respectively. A batch size of 16 was used for all
experiments.



\begin{table}[t]
\centering
\caption{Computational infrastructure for PASCAL VOC experiments.}
\begin{tabular}{lll}
Hardware/Software  & Type/Version \\\hline
CPU     & Intel(R) Xeon(R) Silver \\
        & 4216 CPU @ 2.1GHz \\
GPU model & Tesla V100 Volta 32G \\
GPU driver & 440.33.01 \\ 
OS      & CentOS 7.7.1908 Core \\
Python  & 3.6.3 \\
PyTorch &  1.2.0 \\
CUDA    &  10.2
\end{tabular}
\label{table:comp-architechture-cedar-pascal-voc-appendix}
\end{table}

\begin{table}[t]
\centering
\caption{Search spaces for PASCAL VOC image segmentation.}
\begin{tabular}{lll}
  Hyperparameters  & Default &  Space  \\\hline
  learning rate    & 0.002 & log($10^{-5}$, $10^{-2}$) \\
  momentum         & 0.9 & lin($0.50$, $0.99$) \\
  weight decay     &  0.000001 & log($10^{-8}$, $10^{-1}$) \\
  batch-size       & 16 & -
\end{tabular}
\label{table:hps-pascal-voc-space-appendix}
\end{table}

\subsection{Major histocompatibility class I-associated peptide binding prediction with shallow MLP}

\begin{table}[t]
\centering
\caption{Search spaces for the different hyperparameters for the MLP-MHC task}
\begin{tabular}{rll}
  Hyperparameters      & Default & Space  \\\hline
  hidden layer size         & & lin($20$, $400$) \\
  L2-weight decay         & & log($0$, $1$) 
\end{tabular}
\label{table:hps-space-mhc}
\end{table}

\begin{table}[t]
\centering
\begin{tabular}{rll}
  \# HPs & Hyperparameters      & Default Value  \\
  1 &     hidden layer size         & 150 \\
  2 &     L2-weight decay & 0.001
\end{tabular}
\caption{Defaults for MLP-MHC task.}
\label{table:hps-mhc-default-appendix}
\end{table}

\begin{table}[t]
\centering
\caption{Comparison of performance on datasets}
\begin{tabular}{c|c|c|c|c}
  \hline
  Model name & Dataset & AUC & PCC \\
  \hline
  NetMHCpan4&HPV&0.53 &0.39\\
  MHCflurry& HPV& 0.58 &0.41\\
  MLP-MHC&HPV &0.63 & 0.31\\
  \hline
  \hline
  NetMHCpan4&NetMHC-CVsplits&0.854 &0.620\\
  MHCflurry& NetMHC-CVsplits&0.964* &0.671*\\
  MLP-MHC&NetMHC-CVsplits &0.861 & 0.660\\
  \hline
\end{tabular}
\label{table:mhc-models}
\end{table}

\begin{table*}[t]
\centering
\caption{Comparison of models for the MLP-MHC task}
\begin{tabular}{c|c|c|c|c}
  \hline
  Model name & Inputs& Model design& Dataset & Sequence encoding \\
  \hline
  NetMHCpan4&allele+peptide& shallow MLP& custom CV split\cite{vita2019immune}& BLOSUM62\\
  MHCflurry& peptide& ensemble of shallow MLPs& \cite{o2018mhcflurry}& BLOSUM62\\
  MLP-MHC&allele+peptide& shallow MLP & same as \cite{o2018mhcflurry}& Sparse\\
\end{tabular}
\label{table:mhc-models}
\end{table*}

\begin{table}[t]
\centering
\caption{Computational infrastructure for MLP-MHC experiments.}
\begin{tabular}{lll}
Hardware/Software  & Type/Version \\\hline
CPU     & Intel(R) Xeon(R) CPU E5-2640 v4 \\
        & 320 CPU @ 2.40GHz \\
OS      & CentOS 7.7.1908 Core \\
Python  & 3.6.8 \\
sklearn &  0.22.2.post1 \\
BLAS &  3.4.2 \\
\end{tabular}
\label{table:comp-architechture-mhc}
\end{table}

\paragraph{Task}
The MLP-MHC is a regression task with the goal of predicting the relative binding affinity for a given peptide and major histocompatibility complex class I (MHC) allele pair.
The major histocompatibility complex (MHC) class I proteins are present on the surface of most nucleated cells in jawed vertebrates \cite{pearson2016mhc}.
These proteins bind short peptides that arise from the degradation of intra-cellular proteins \cite{pearson2016mhc}.
The complex of peptide-MHC molecule is used by immune cells to recognize healthy cells and eliminate cancerous or infected cells, a mechanism studied in the development of immunotherapy and vaccines \cite{o2018mhcflurry}.
The peptide binding prediction task is therefore at the base of the search for good vaccine and immunotherapy targets \cite{o2018mhcflurry,jurtz2017netmhcpan}.

The input data is the concatenated pairs of sequences: the MHC allele and the peptide sequence.
For the MHC alleles, we restricted the sequences to the binding pocket of the peptide, as seen in \cite{jurtz2017netmhcpan}.
The prediction target is a normalized binding affinity score, as described in \cite{jurtz2017netmhcpan,o2018mhcflurry}.

\paragraph{Datasets and sequence encoding}

While both \emph{MHCflurry} and \emph{NetMHCpan4} models use a BLOSUM62 encoding \cite{henikoff1992amino} for the amino acids, in we chose to instead encode the amino acids as one-hot as described in \cite{nielsen2007netmhcpan}.

The \emph{NetMHCpan4} model is trained on a manually filtered dataset from the immune epitope database \cite{vita2019immune,jurtz2017netmhcpan} that has been split into five folds used for cross-validation, available on the author's website \cite{jurtz2017netmhcpan}.

In contrast, the \emph{MHCflurry} model is trained on a custom multi-source dataset (available from \href{http://dx.doi.org/10.17632/8pz43nvvxh.3#file-5a2034f4-aa0f-4622-9e89-b35c765fb6ce}{Mendeley data} and the \cite{o2018mhcflurry} publication cite) and validated/tested on two external datasets from \cite{pearson2016mhc} and an HPV peptide dataset available at the same website as above.

\paragraph{Bootstrapping} We have three different sets for training, validating and testing.
We thus performed bootstrapping separately on each set for every training and evaluation.

\paragraph{Model}
The model is a shallow MLP with one hidden layer from \emph{sklearn}. 
We used the default setting for the non-linearity \emph{relu} and weight initialization strategy (\cite{pmlr-v9-glorot10a}).
The following table (Table \ref{table:mhc-models}) offers some comparison points between our model and the \emph{NetMHCpan4} \cite{jurtz2017netmhcpan} and \emph{MHCflurry} \cite{o2018mhcflurry} models.

While the \emph{MHCflurry} model \citep{o2018mhcflurry} train only no the peptide sequences and uses ensembling to perform its predictions, training multiple models for each MHC allele, the \emph{NetMHCpan4} model \citep{jurtz2017netmhcpan} uses the allele sequence as input and trains one single model. 

We chose to retain the strategy proposed by the \emph{NetMHCpan4} model, where a single model is trained for all alleles \cite{jurtz2017netmhcpan}.
As a reference, MHCflurry uses ensembling to perform predictions; indeed, the authors report that for each MHC allele, an ensemble of 8-16 are selected from the 320 that were trained \cite{o2018mhcflurry}. 

\paragraph{Search space of hyperparameters}
For the hyperparameter search, we selected hidden layer sizes between 20 and 400 (Table \ref{table:hps-space-mhc}), to engulph a range slightly larger than the ones described by both \cite{jurtz2017netmhcpan,o2018mhcflurry}.
The second hyperparameter that was explored was the L2 regularisation parameter, for which a log-uniform range between 0 and 1 was explored.

\paragraph{Comparison of performance}
We would like to state the goal of the present study was not to establish new state of the art (SOTA) on the MHC-peptide binding prediction task.
However, we still report that when comparing the performance of our model to those of \emph{NetMHCpan4} and \emph{MHCflurry} we found the performance of our model comparable. 
Briefly, for the results in Table \ref{table:mhc-models}, we used the existing pre-trained \emph{NetMHCpan4} and \emph{MHCflurry} tools to predict the binding affinity of both datasets: the previously described HPV external test data (HPV) from \cite{o2018mhcflurry} and the cross-validation test datasets from \cite{jurtz2017netmhcpan} (NetMHC-CVsplits).

We would like to point out that since the \emph{MHCflurry} model was published later than the \emph{NetMHCpan4} one, there is a high chance that the dataset from the cross-validation splits (NetMHC-CVsplits) may be contained in the dataset used to train the existing \emph{MHCflurry} tool.
The proper way to compare performances would be to re-train the \emph{MHCflurry} model on each fold and test susequently its performance; however, since our goal is not to reach new SOTA on this task, we leave this experiment to be performed at a later time.

This would result in a likely overestimation of the performance of \emph{MHCflurry} on this dataset, which we noted with the $*$ sign in Table \ref{table:mhc-models}.

A more in-depth study is necessary to compare in a more through way this performance with respect to the differences in model design, dataset encoding and other factors.


\section{Hyperparameter optimization algorithms}
\label{sec:hpo-methods-appendix}

\subsection{Grid Search}
Let $a_i$, $b_i$ and $n$ be the hyperparameters of the grid search, where $a_i$ and $b_i$ are 
vectors of minima and maxima for each dimension of the search space, and 
$n$ is the number of values per dimension. 
We define $\Delta_i$ as the interval between
each value on dimension $i$. A point on the grid is defined by $p_{ij}=a_i + \Delta_i(j-1)$.
Grid search is simply the evaluation of $r(\lambda)$
from Equation~\ref{eq:hpo_loss} on all possible combinations of values $p_{ij}$.

\subsection{Noisy Grid Search}
\label{sec:noisy-grid-search}

Grid search is a fully deterministic algorithm. Yet, it is highly sensitive to the 
design of the grid. To provide a variance estimate of similar choices of the grid 
and to be able to distinguish lucky grid, we consider a noisy version of grid search.

For the noisy grid search,
we replace $a_i$ by $\tilde{a}_i \sim U(a_i - \frac{\Delta_i}{2}, a_i + \frac{\Delta_i}{2})$ and similarly for $b_i$.
$\tilde{\Delta}_i$ and $\tilde{p_{ij}}$ then follows from $\tilde{a}_i$ and $\tilde{b}_i$.
In expectation, noisy grid search will cover the same grid as grid search, as proven below.

\begin{align*}
  \E[\tilde{p}_{ij}] 
      & = \Ee{\tilde{a}_i + \tilde{\Delta}_i(j-1)}  \\
      & = \Ee{\tilde{a}_i + \frac{\tilde{b} - \tilde{a}}{n-1}(j-1)} \\
      & = \Ee{\tilde{a}_i} + \Ee{\frac{\tilde{b}}{n-1}}(j-1)- \Ee{\frac{\tilde{a}}{n-1}}(j-1) \\
      & = a_i + \frac{b}{n-1}(j-1)- \frac{a}{n-1}(j-1) \\
      & = a_i + \frac{b-a}{n-1}(j-1) \\
      & = a_i + \Delta_i(j-1) = p_{ij}
\end{align*}

This provides us a variance estimate of grid search that we can compare against non-deterministic
hyperparameter optimization algorithms.

\subsection{Random Search}

The search space of random search will be increased by 
$\pm \frac{\Delta_i}{2}$ as defined for the noisy grid search to ensure that they both cover the
same search space. 
For all hyperparameters, the values are sampled from a uniform
$p_i \sim U(a_i - \frac{\Delta_i}{2}, b_i + \frac{\Delta_i}{2})$. For learning rate and weight
decay, values are sampled uniformly in the logarithmic space.

%
%

%
%

\section{Hyperparameter optimization results}
\label{sec:hpo-methods-results}

\begin{figure}
  \center
  \includegraphics[width=\columnwidth]{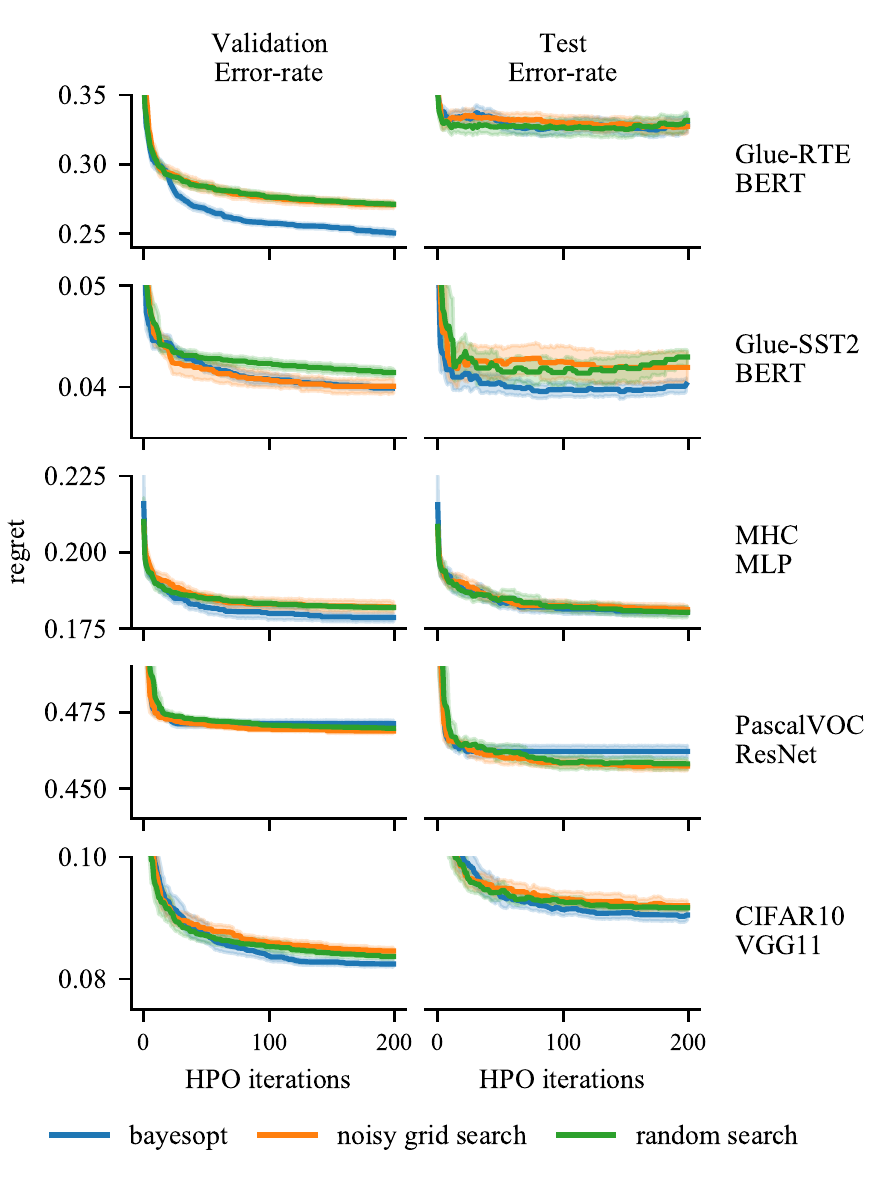}
    \caption{
      \textbf{Optimization curves of hyperparameter optimization executions}
      Each row presents the result for a different task. Left column are results
      on validation set, the one hyperparameters were optimized on. Right column
      are results on the test sets. Hyperparameter optimization methods are Bayesian Optimization,
      Noisy Grid Search (See Section~\ref{sec:noisy-grid-search}), and Random Search.
      The y-axis are the best objectives found until an iteration $i$, on a different scale for each
      task. Left and right plots share the same scale on y-axis, so that we can easily observe
      whether validation error-rate corresponds to test error-rate. The bold lines are averages and
      the size of lighter colored areas represents the standard deviations. They are computed based
      on 20 independent executions for each algorithms, during which only the seed
      of the hyperparameter optimization is randomized. For more details on the experiments see
      Section~\ref{sec:exp_var_hpo}. Two striking results emerge from these graphs. 1) The typical
      search spaces are well optimized by all algorithms, and in some cases there is even signs of 
      slight over-fitting (on BERT tasks). 2) The standard deviation stabilizes early, before 50
      iterations in most cases. These results suggests that larger budgets for hyperparameter
      optimization would not reduce the variability of the results in similar search spaces.
      This is likely not the case however for more complex search spaces such as those observed in the 
      neural architecture search literature.
  }
  \label{fig:hpo-curves}
\end{figure}

Figure~\ref{fig:hpo-curves} presents the optimization curves of the hyperparameter optimization
executions in Section~\ref{sec:exps-var}.

\section{Normality of performance distributions in the case studies}

\begin{figure}
  \center
  \includegraphics[width=\columnwidth]{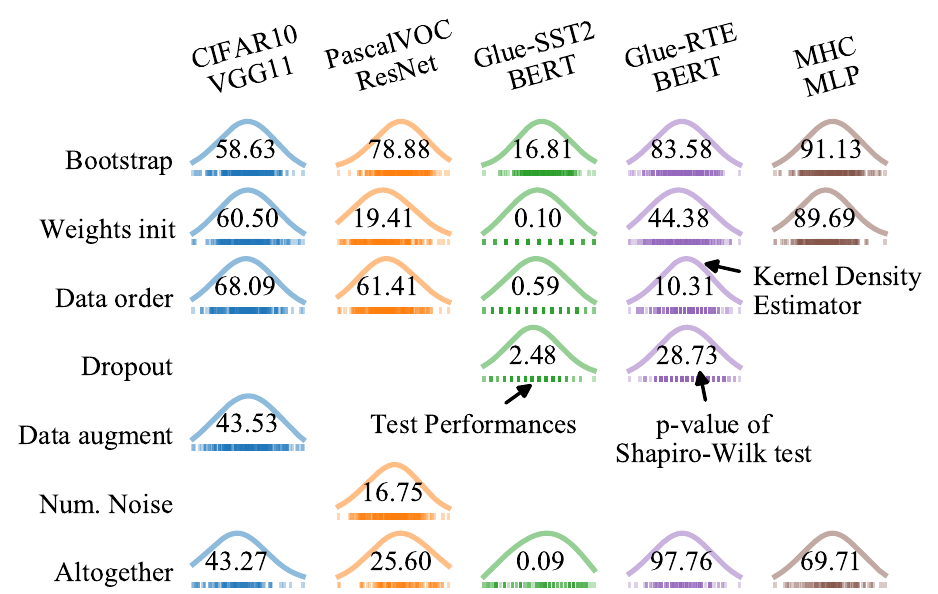}
    \caption{
      \textbf{Performance distributions conditional to different sources of variations.}
      Each row is a different source of variation. For each source, all other
      sources are kept fix when training and evaluating models.
      The last row presents the distributions when all the sources of variation
      are randomized altogether.
      Each column is the results for the different tasks.
      We can see that except for Glue-SST2 BERT, all case studies have distributions
      of performances very close to normal. In the case of Glue-SST2 BERT, we note that
      the size of the test set is so small that it discretizes the possible performances.
      The distribution is nevertheless roughly symmetrical and thus amenable to many
      statistical tests.
  }
  \label{fig:normal}
\end{figure}

Figure~\ref{fig:normal} presents the Shapiro-Wilk test of normality on all our results
on sources of variations.

\section{Randomizing more sources of variance increase the quality of the estimator}

\begin{figure}
  \center
  \includegraphics[width=\columnwidth]{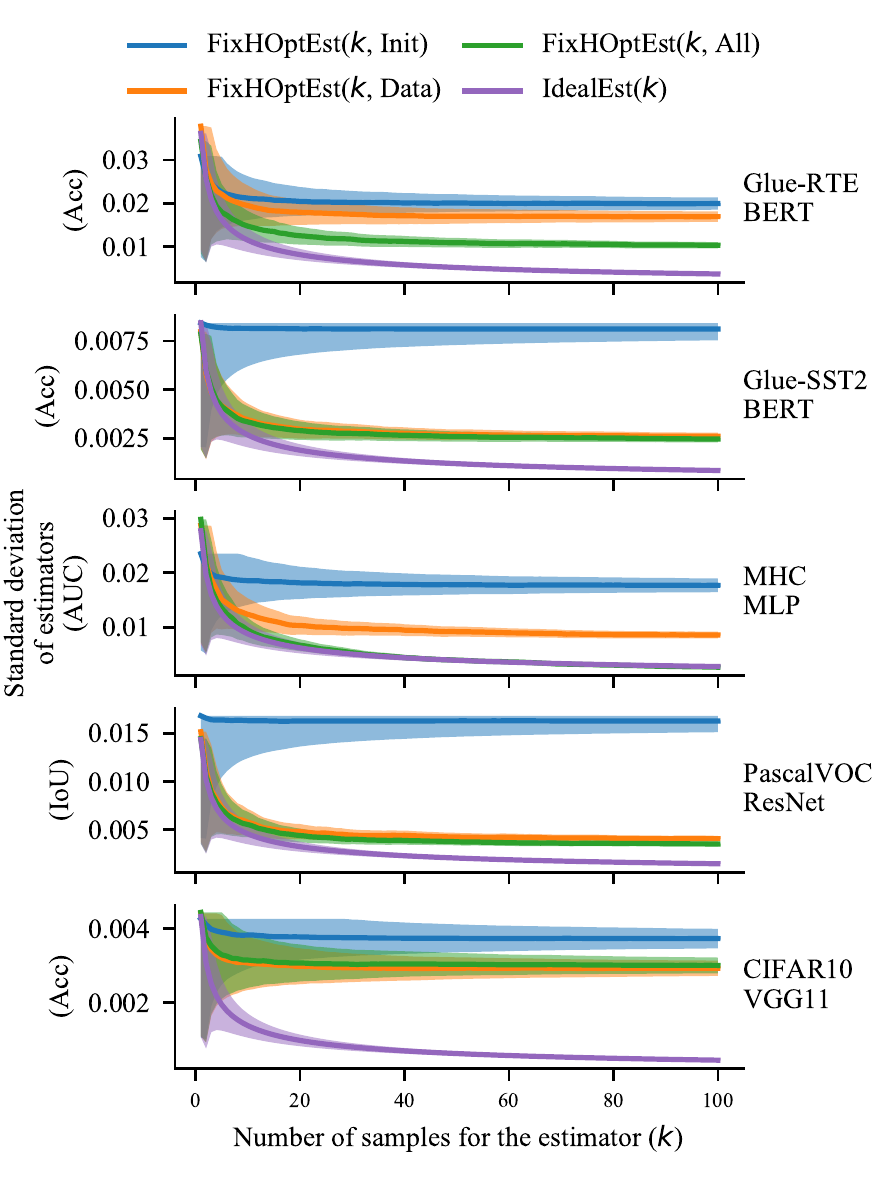}
    \caption{
      \textbf{Standard error of biased and ideal estimators with $k$ samples.}
      Each plot represents the standard error of the different tasks described in
      Section~\ref{sec:exps-var}.
    On x axis, the number of samples used by the estimators to compute the average performance.
    On y axis, the standard deviation of the estimations, in terms of task objective;
    Classification accuracy (Acc),
    Intersection over Union (IoU),
    Area Under the Curve (AUC).
    Uncertainty represented in light color is computed analytically as the approximate standard
    deviation of the standard deviation of a normal distribution computed on $k$ samples.
    For all case studies,
    \textbf{accounting for more sources of variation reduces or keeps constant the 
      standard error of $\hmuk$.}
      In all case studies, only accounting for weights initialization,
    \texttt{FixHOptEst(k, Init)}, is by far the worst estimator.
    Comparatively, \texttt{FixHOptEst(k, All)} provides a systematic improvement
    towards \texttt{IdealEst(k)} for no additional computational cost compared to
    \texttt{FixHOptEst(k, Init)}.
    \textbf{Ignoring variance from $\Hopt$ is harmful for a good estimation of $\expr$.}
    The MHC task with MLP is the only one for which \texttt{FixHOptEst(k, All)}
    matches \texttt{IdealEST(k, All)}. We suspect this may be explained by the 
    relatively small standard deviation due to hyperparameter optimization observed in
    Figure~\ref{fig:variances}. \texttt{FixHOptEst(k, All)} would have thus captured most of the
    variability in the learning pipeline.
        }
    \label{fig:standard-error-all}
\end{figure}

\begin{figure}
  \center
  \includegraphics[width=\columnwidth]{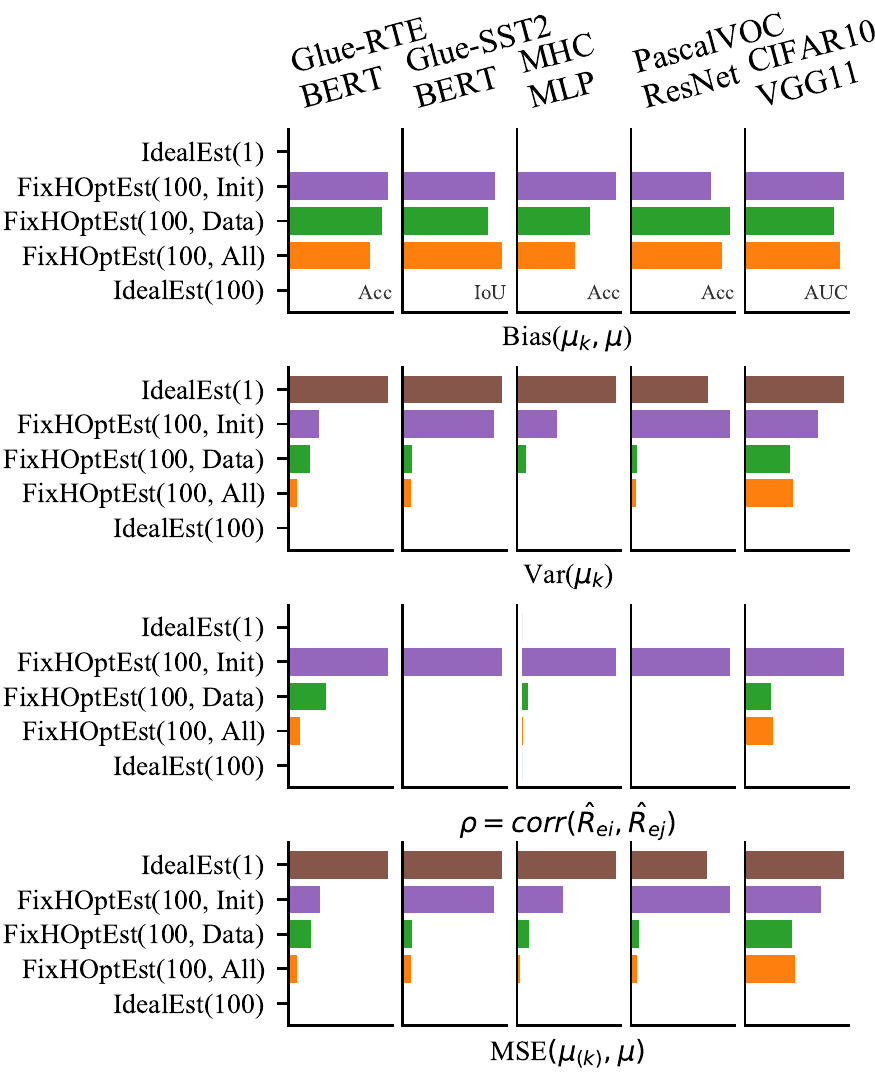}
    \caption{
      \textbf{Decomposition of the Mean-Squared-Error for different estimators
       of $\expr$.}
       On each sub-figure from top to bottom, 1) bias between the estimator
       and the expected empirical risk $\mbox{Bias}(\mu_{(k)}, \mu)$,
       2) variance of the estimator $\Var(\mu_{(k)})$,
       3) correlation
       between performances measures $\hat{R}_e$ as presented
       in Equation~\ref{eq:tmuk-var} and
       4) the mean-squared-error of the estimator $\mbox{MSE}(\mu_{(k)}, \mu)$.
       For each sub-figure, each row 
       is a different estimators, with \texttt{IdealEst(k=1)} as a comparison point.
       The experimental procedure to compute these statistics are described in section
       \autoref{sec:hopt-variance}.
       Without any surprise the \texttt{IdealEst(100, All)} minimizes the mean-squared-error
       so well that it looks close to 0 on the figure compared to the other estimators.
       Among the other estimators, the mean-squared-error is reduced most significantly by 
       \texttt{FixedHOptEst(100, All)} on all tasks.
       If we look at the decomposition of the mean-squared-error, i.e., the bias
       and the variance, we see on first sub-figure that the bias is stable across all biased
       estimators on all tasks, while on second sub-figure the variance varies widely. It is thus
       the reduced variance of the biased estimators that leads to improved mean-squared-error. This
       is a counter-intuitive result because the estimator with lowest variance are these
       accounting for more sources of variations. The intuition is thus that they should have more
       variance, not less. We derived the variance of the biased estimators in 
       Equation~\ref{eq:tmuk-var} which highlighted that the correlation among performances 
       $\hre$ can increase the variance of the biased estimators. We can see in the third
       sub-figure that this correlation drastically drops when accounting for more 
       sources of variances. The mean-squared-error, in other words the quality of the estimators,
       is thus significantly improved by decorrelating the performance measures.
    }
    \label{fig:mse-decomposition}
\end{figure}

Figure~\ref{fig:standard-error} only presented the Glue-RTE and CIFAR10 tasks.
We provide here a complete picture of
the standard deviation of the different estimators in
Figure~\ref{fig:standard-error-all}.
We further present a decomposition of the mean-squared-error in
Figure~\ref{fig:mse-decomposition}
to help understand why accounting for more sources of
variations improves the mean-squared-error of the biased estimators.

\section{Analysis of robustness of comparison methods}
\label{sec:test-analysis}

\begin{figure*}
  \center
  \includegraphics[width=\textwidth]{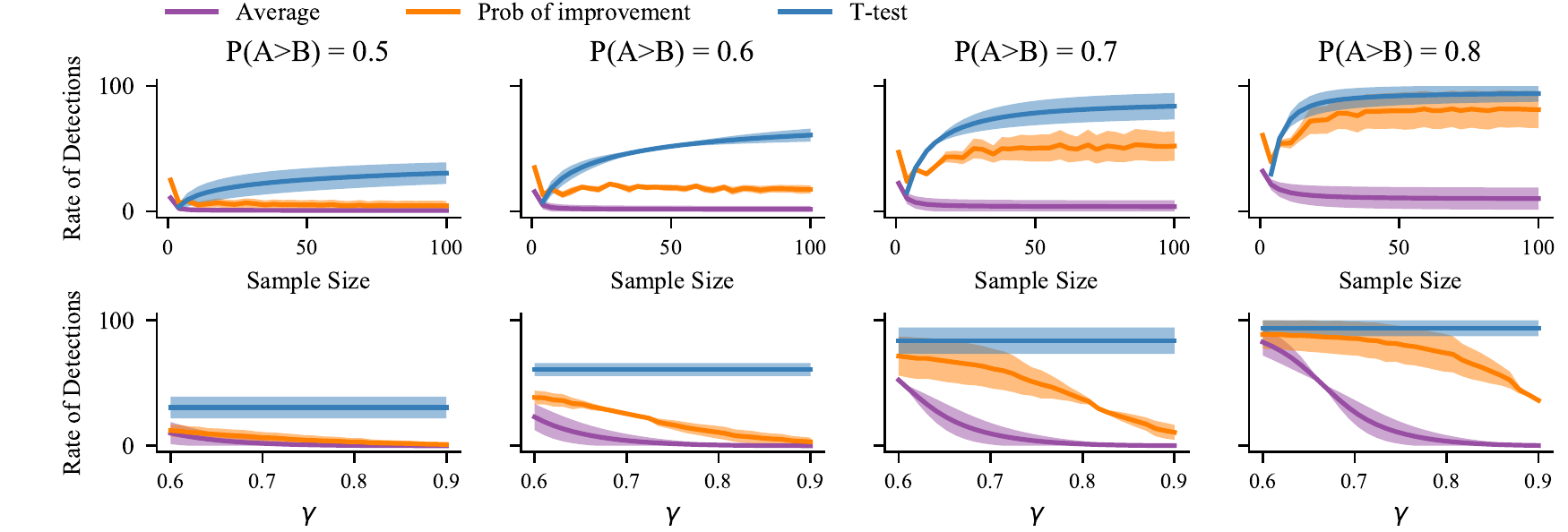}
    \caption{
      \textbf{Analysis of the robustness of comparison methods.}
      On the first row, rate of detections of comparison methods in function
      of the sample size. On the second row, rate of detections
      of comparison methods in function of the threshold $\gamma$.
      Each column are simulations with different true simulated
      probability of 
      of a learning algorithm $A$ to outperform another algorithm 
      $B$ across random fluctuations (ex: random data splits).
    }
  \label{fig:test-analysis}
\end{figure*}

In addition to simulations described in Section~\ref{sec:error_rates}, 
we executed experiments in which we varied the sample size and the 
threshold $\gamma$. To select the threshold of the average, we converted
$\gamma$ into the equivalent performance difference $(\delta=\Phi^{-1}(\gamma)\sigma)$.
Results are presented in Figure~\ref{fig:test-analysis}

\end{document}